\documentclass[letterpaper, 10 pt, conference]{ieeeconf}  

\IEEEoverridecommandlockouts                              

\usepackage{graphics} 
\usepackage{epsfig} 
\usepackage{mathptmx} 
\usepackage{times} 
\usepackage{amsmath} 
\usepackage{amssymb}  
\usepackage{mathtools}
\usepackage{algorithmic, algorithm, threeparttable, type1cm}
\usepackage{graphicx, textcomp, xcolor}
\usepackage{fancyhdr}
\usepackage{hyperref}
\fancypagestyle{firststyle}
{
   \fancyhf{}
   \fancyhead[C]{{\large Preprint:}} 
}

\title{\LARGE \bf
Distributed Algorithm with Emergent Area Partitioning and Base Station's Situation Awareness for Multi-Robot Patrolling
}

\author{Kazuho Kobayashi$^{1}$, Shohei Kobayashi$^{1}$, Seiya Ueno$^{2}$, and Takehiro Higuchi$^{2}$
\thanks{$^{1}$Kazuho Kobayashi and Shohei Kobayashi is with Department of Mechanical Engineering, Materials Science, and Ocean Engineering, Graduate School of Engineering Science, Yokohama National University, Yokohama-city, Kanagawa, Japan
        {\tt\small kobayashi-kazuho-dj@ynu.jp \textit{or} kazuho.kobayashi.ynu@gmail.com}}%
\thanks{$^{2}$Seiya Ueno and Takehiro Higuchi with Faculty of Environment and Information Sciences, Yokohama National University
        {\tt\small \{ueno-seiya-wk, higuchi\}@ynu.ac.jp}}%
\\
\textbf{An open-access published version of this paper is available at DOI: \href{https://doi.org/10.20965/jrm.2025.p0927}{https://doi.org/10.20965/jrm.2025.p0927}.}}

\begin{document}
\maketitle
\thispagestyle{empty}
\pagestyle{empty}

\thispagestyle{firststyle} 
\begin{abstract}
Patrolling with multiple robots offers efficient surveillance to detect and manage undesired situations. This necessitates improved patrol efficiency and operator situation awareness at base stations. Enhanced situation awareness enables operators to predict robots' behaviors, support recognition and decision-making, and execute emergency interventions. This study presents the Local Reactive and Partition (LR-PT) algorithm, a novel multi-robot patrolling approach. In simulations, LR-PT outperformed existing methods by ensuring frequent patrols of all locations of interest and enhancing the situation awareness of the base station. Robots independently select patrol targets based on locally available information, integrating patrol needs and the urgency of reporting mission progress to the base station into a unified utility function. This locality also contributes to robustness against communication constraints and robot failures, as demonstrated in this research. The algorithm further autonomously emerged the area partition, which can avoid falling into local optima and realize the comprehensive patrol over the whole mission area. The simulation results demonstrated the superior performance of LR-PT for multi-robot patrolling, utilizing the advantages of swarm robotics and addressing real-world operational challenges.
\end{abstract}

\section{INTRODUCTION}
Patrolling represents a promising application for multi-robot systems or robotic swarms aimed at early detection and reporting of undesired situations. Multiple robots can be widely spread to survey various locations and detect targets earlier~\cite{Alotaibi2019}. Such systems are also robust, flexible, and expandable to various mission configurations~\cite{Sahin2005}. 

Despite these advantages, real-world applications of multi-robot systems remain limited~\cite{Schranz2020}. One of the reasons is their unpredictable behavior: human operators cannot predict the systems' actions, outcomes, and potential side effects. Multi-robot systems often utilize local interactions among robots, which are usually invisible and thus reduce predictability. Systems deployed remotely from human operators may further escalate this challenge. Resultantly, the lower predictability for humans leads to concerns about harmful behaviors and delays in addressing such behaviors. In particular, for life-threatening or complicated missions~\cite{Carrillo-Zapata2020,Watanobe2023,Branesh2024}, relieving the concerns is a key to promoting real-world applications.

One solution is continuous or frequent network connectivity between robots and a base station (BS). This connectivity allows robots to report mission progress and their status, enhancing the situation awareness (SA)~\cite{Endsley1988} of the BS. Improved SA enhances predictability by better understanding about mission progress and early detection of system malfunctions. Furthermore, the operators can intervene in the systems to support robots' recognition and high-level decision-making~\cite{Kolling2016}. In these contexts, the BS serves as the interface for operators to supervise and make high-level decisions. 

This research proposes the Local Reactive and Partition (LR-PT) algorithm for multi-robot patrolling. The research contributes to the area of multi-robot patrolling by:
\begin{itemize}
  \item Developing a distributed patrol algorithm that balances patrolling and recovery of connectivity to the BS.
  \item Evaluating the robustness and scalability of the algorithm as well as its performance.
\end{itemize}
Existing patrol algorithms that emphasize connectivity frequently depend on centralized coordination or strict connectivity constraints. This study expands on those efforts by introducing a distributed mechanism that considers both efficient patrolling and BS's SA. The study builds upon the preliminary research in \cite{Kobayashi2023arxiv}, introducing an expanded algorithm incorporating emergent area partitioning and further evaluations. The following sections describe the related works (Section 2), problem settings (Section 3), proposed algorithm (Section 4), simulation studies (Section 5), and conclusions (Section 6).

\section{RELATED WORKS}
While numerous distributed algorithms have been proposed, few specifically address the connectivity to BSs. For instance, Machado et al. introduced a Conscientious Reactive algorithm, showing superior performance over some centrally organized algorithms~\cite{Machado2003}. Ant-based algorithms inspired by insect behavior have been explored by Glad et al.~\cite{Glad2008,Glad2009}. More recently, Portugal et al. developed Bayesian-inspired algorithms~\cite{Portugal2013a,Portugal2016} that allow robots to update parameters indicating the suitability of patrolling specific locations. Othmani-Guibourg et al. adopted the learning approach called the Long Short Term Memory (LSTM) strategy to emulate the centralized algorithm distributedly~\cite{Othmani-Guibourg2018, Othmani-Guibourg2019}. Yan et al. introduced the Expected Reactive algorithm, which considers patrol utility: the patrol need of a location and travel cost to the location~\cite{Yan2016}. This utility-based approach has also been incorporated into auction-based algorithms to allocate the location to the most suitable robot. Though many auction-based patrolling depend on the centrally facilitated processes (e.g.,~\cite{Hwang2009,Pippin2013}), Farinelli et al. developed the Dynamic Task Assignment (DTA) method, which considers patrol needs, travel cost, and appropriate assignment based on auctions in a distributed manner~\cite{Farinelli2017}. As a simpler approach in some aspects, Sato et al. developed a probabilistic behavior model to allocate patrol areas for swarm robots~\cite{Sato2023}. Building on these distributed methods, which share similar problem settings and algorithm ideas, this study introduces an additional concept to improve mission predictability by defining the BS and incorporating a mechanism to consider its SA.

Several examples consider the connectivity to BSs, though they are usually centrally coordinated. Scherer and Rinner devised the centralized method, which ensures connectivity to the BS and latency constraints~\cite{Scherer2017,Scherer2020}. They also developed algorithms for persistent surveillance with continuous connectivity, though they are also centrally coordinated~\cite{Scherer2016,Scherer2020sh}. Banfi et al. developed a centralized algorithm to minimize the communication latency during patrolling missions~\cite{Banfi2015}. With slight differences in mission requirements, multi-robot exploration algorithms that maintain inter-robot communication connections were also developed~\cite{Amigoni2017,Nestmeyer2017}. Yanmaz et al. developed and compared two algorithms that prioritize coverage and connectivity, respectively~\cite{Adam2019}. Kobayashi et al. proposed the distributed patrol algorithm with continuous connectivity~\cite{Kobayashi_2024jrm}. Though they do not necessarily focus on patrol missions, numerous studies also consider connectivity constraints, which can be utilized to provide situation awareness~\cite{Hung2020, Murayama2023,Sano2023,Caregnato-Neto2023}. Some of the aforementioned efforts depend on centralized coordination to maintain or recover connectivity to the BS. Furthermore, continuous connectivity may excessively restrict the patrol area or require many robots to be deployed to reach remote locations in mission areas.

This study expands these existing efforts by introducing an SA-aware algorithm, while relieving constraints on operational range and requirements for centralized planning. These improvements are implemented with robustness and scalability, utilizing fully-distributed behavior. The following sections show details of the proposed algorithm: Local Reactive and Partition (LR-PT).

\section{PROBLEM DEFINITION}\label{Prob}
The problem settings of this study are common to the existing research~\cite{Kobayashi_2024jrm}. The field corresponds to a large open field, such as a disaster zone, border region, or sea surface. Further details can be found in~\cite{Kobayashi_2024jrm}. These settings are also virtually equivalent to many previous studies that modeled environments as graph maps with known nodes and edges (e.g., \cite{Machado2003,Yan2016,Farinelli2017}). The grid map defined in the following subsection generalizes graph maps, particularly complete graphs, and is especially suited for modeling and visualizing large open fields.

\subsection{Field and Mission}
The patrol area is modeled as a two-dimensional grid map, defined as $G \coloneqq \{g^k\ |\ k=1,2,\ldots,K\}$, where each grid $g^k$ represents a location that requires patrolling. $k$ identifies each grid, and $K$ is the total number of grids. Each grid is square, and the location of its center represents $g^k$'s location: $\textit{\textbf{x}}^k$. A patrol for $g^k$ is considered completed when a robot approaches within $\rho$ [m] from $\textit{\textbf{x}}^k$. The coordinate system follows the article \cite{Kobayashi_2024jrm}. The map does not include obstacles since the research assumes an expansive open area.

\begin{figure}[t]
  \centering
  \resizebox*{5cm}{!}{\includegraphics{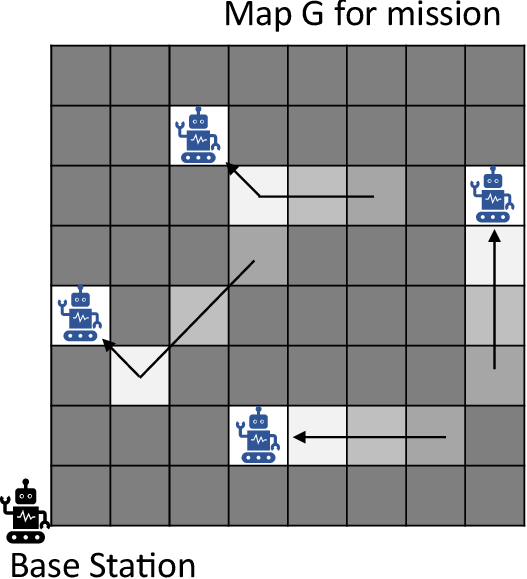}}
  \caption{A conceptual image of a patrol mission with five robots in a field with $K=64\ (8 \times 8)$ grids. The grid color indicates its idleness; the darker the color, the higher the idleness. The figure is reproduced from Figure 1 in \cite{Kobayashi2023robomech} with modified captions.}
  \label{fig_concept}
\end{figure}

Patrolling is conceptualized as a series of sequential visits to grids by each robot, with the goal of minimizing the time between visits to each grid. This temporal dimension is encapsulated by the concept of grid idleness~\cite{Machado2003} denoted as $i^k(t)$. It is defined as the time elapsed since a grid was last patrolled. For instance, when a robot patrolled $g^k$ at $t^k$ and no other robots have patrolled $g^k$ since then, the idleness of the grid at current time $t$ can be computed as $i^k(t) = t - t^k$. In general, the primary objective of the patrolling mission is to keep $i^k(t)$ as low as possible. Mission times $t$ are timesteps corresponding to seconds, which take non-negative integer numbers: $t_0 \leq t \leq T$. $t_0$ is when the mission initiates, and $T$ is when the mission terminates. Figure~\ref{fig_concept} shows the conceptual image of such a mission. In the figure, the grid color indicates its idleness; the darker the color, the higher the idleness.

\subsection{Robots}
This study considers robots on a two-dimensional plane, such as ground vehicles, autonomous boats, or quad-rotor drones operating at a constant altitude. First, this study defines robots as $R \coloneqq \{r_n\ |\ n=1,2,\dots,N\}$. Here, $r_n$ is a robot whose ID is $n$, and $N$ is the total number of robots. $r_1$ serves as a BS to interface to human operators, and $r_2, \dots, r_N$ carry out patrolling with homogeneous capabilities. Other assumptions about the robots' capabilities, such as self-localization of their position $\textit{\textbf{x}}_n(t)$ and orientation $\theta_n(t)$, sensor range $r_s$ to identify each other, and communication range $r_c(>r_s)$ to exchange messages with $r_n$'s communication neighbor $A_n(t)$, are consistent with those in the article \cite{Kobayashi_2024jrm}. 

Regarding behavior during missions, each robot $r_n$ selects its patrol target grid $g_n^{\tau}$ upon arriving at its current target grid. The following section describes the selection algorithm as the main part of the proposed algorithm. The article~\cite{Kobayashi_2024jrm} details the motion model of each robot to navigate the robot to its patrol target grid.

\subsection{Performance Metrics}
This research introduces two perspectives for evaluating mission performance: patrolling performance, and BS's SA. The former perspective, quantified by idleness, represents the frequency and comprehensiveness with which the robots patrolled the designated map. On the other hand, the latter perspective reflects the effectiveness of the robots in providing SA to the BS. The following paragraphs provide details of each metric, while the article \cite{Kobayashi_2024jrm} may support the descriptions.

Two metrics, \textit{graph idleness}: $I_G$ and \textit{worst idleness}: $I_W$ were introduced to evaluate mission performance from the perspective of idleness~\cite{Machado2003}. At a given time instance $t$, an average value of idleness over all grids can be computed as instantaneous graph idleness: $I_g(t)$ by:
\begin{equation}
    I_g(t) = \frac{1}{K}\sum_{k=1}^{K}i^k(t)   
\end{equation}
and graph idleness $I_G$ represents the average of visit intervals across the entire mission time and all grids, defined as:
\begin{equation}
    I_G = \frac{1}{T}\sum_{t=t_0}^{T}I_g(t)
\end{equation}
On the other hand, worst idleness $I_W$ represents the maximum value of idleness across all grids, indicating whether a particular grid has been left unpatrolled for longer than others. The larger value of this metric highlights the potential neglect of malfunctioning instruments or victims for a long time. The instantaneous worst idleness $I_w(t)$ can be computed by:
\begin{equation}
    I_w(t) = \mathop{\mathrm{max}}_{k \in K}i^k(t)
\end{equation}
and $I_W$ can be computed as:
\begin{equation}\label{eq_IW}
    I_W = \mathop{\mathrm{max}}_{t_0 \leq t\leq T}I_w(t)
\end{equation}

For the SA metrics, this study introduces \textit{mean SA delay} and \textit{worst SA delay}~\cite{Kobayashi_2024jrm}. The grid SA delay $D_{\mathrm{SA}}^k(t)$ for $g^k$ denotes the delay of information about $g^k$ held by BS at time $t$. This value increases over time: $D_{\mathrm{SA}}^k(t) = D_{\mathrm{SA}}^k(t-1) + 1$ as long as no robots report information about $g^k$ to the BS, even if they have already patrolled the grid. When a robot acquires the information about $g^k$ at time $t_1$ (i.e., patrolled $g^k$ at $t_1$), and reports the fact to the BS, the BS updates the grid SA delay as $D_{\mathrm{SA}}^k(t) = t-t_1$. The value corresponds to the interval of deliveries for images, sample specimens, or other information in real-world missions. Based on this definition, \textit{instantaneous mean SA delay} $D_{\mathrm{mSA}}$ is the average value of $D_{\mathrm{SA}}^k(t)$ over all grids at $t$:
\begin{equation}\label{eq_instDmSA}
    D_{\mathrm{mSA}}(t) = \frac{1}{K}\sum_{k=1}^{K}D_{\mathrm{SA}}^k(t) 
\end{equation}
and $D_{\mathrm{MSA}}$ over the entire mission is:
\begin{equation}\label{eq_DMSA}
    D_{\mathrm{MSA}} = \frac{1}{T}\sum_{t=t_0}^{T}D_{\mathrm{mSA}}(t) 
\end{equation}
Similar to $I_W$, instantaneous worst SA delay $D_{\mathrm{wSA}}$ is the maximum value of $D_{\mathrm{SA}}^k(t)$ over all grids at $t$:
\begin{equation}\label{eq_instDwSA}
    D_{\mathrm{wSA}}(t)=\mathop{\mathrm{max}}_{k \in K}D_{\mathrm{SA}}^k(t)
\end{equation}
and $D_{\mathrm{WSA}}$ of the mission is:
\begin{equation}\label{eq_DWSA}
    D_{\mathrm{WSA}}=\mathop{\mathrm{max}}_{t_0 \leq t\leq T}D_{\mathrm{wSA}}(t)
\end{equation}
$D_{\mathrm{WSA}}$ highlights the worst-case scenario of delay in which the BS cannot be aware of the information of a specific grid.

For scalability and robustness evaluation, the normalization of metric~\cite{Machado2003} is introduced as follows: 
\begin{equation}\label{eq_norm}
    \overline{M} = M\frac{N-1}{K}
\end{equation}
M is a metric to be normalized, and the overline indicates normalization (e.g., $\overline{I}_G$ for normalized graph idleness). Unlike the original, it multiplies by $N-1$ since the BS does not participate in patrolling. This normalization adjusts for the effects of map size $K$ and swarm size $N$. A constant normalized metric value indicates that the evaluated algorithm is scalable against changes in swarm and map sizes.

\section{PROPOSED METHOD}\label{Algo}
This chapter describes the proposed algorithm Local Reactive and Partition (LR-PT). Each robot executes two behaviors: knowledge update (Section 4.1) and patrol target grid selection (Section 4.2).

\subsection{Knowledge Update}
Each robot $r_n$ maintains \textit{assumed idleness}: $H_n(t)\coloneqq\{h_n^k(t)=[i_n^k(t), t_n^k(t)]\ |\ k=1,2,\ldots,K\}$ as its knowledge about the map $G$. Here, $i_n^k(t)$ represents the assumed value of grid $g^k$'s idleness: $i^k(t)$ by $r_n$. $t_n^k(t)$ denotes the latest update time of $i_n^k(t)$. Since direct observation of $i^k(t)$ via sensors is impractical, each robot bases its behavior on $H_n(t)$. The robots maintain their assumptions based on their patrol outcomes and information from other robots.

Algorithm~\ref{alg_udassume} outlines the procedure for updating assumed idleness, closely following the method in~\cite{Farinelli2017}. Initially, $r_n$ increments its assumed idleness values as time progresses (lines 3-4). It keeps the corresponding $t_n^k(t)$ constant, as the increase in $i^k(t)$ is not verified by actual patrol. Upon completing the patrol of grid $g_n^{\tau}$, $r_n$ updates its assumption $h_n^{\tau}(t)$ (lines 5-6). The grid idleness $i^k(t)$ is reset to zero at this moment by its definition, and the update reflects this change. Subsequently, $r_n$ selects its new patrol target grid (line 7), as outlined in the next section. At the same time, $r_n$ broadcasts $H_n^s(t)$ to all directly connected robots to synchronize their assumptions (lines 8-10). $H_n^s(t)$ is a subset of $H_n(t)$ consisting of $h_n(t)$ entries with the $s$ largest $t_n^k(t)$ values. The value of $s$ reflects the communication bandwidth. Finally, upon receiving $H_m^s(t-1)$ from $r_m \in A_n(t-1)$, $r_n$ updates its knowledge with the most recent information (lines 11-14).

\begin{algorithm}[tbp]
\caption{Update assumption $H_n(t)$ by robot $r_n$ at every timestep}
\label{alg_udassume}
\begin{algorithmic}[1]
    \STATE $H_n(t) \leftarrow H_n(t-1)$
    \FORALL{$i^k_n(t) \in H_n(t)$}
      \STATE $i^k_n(t) \leftarrow i^k_n(t) + 1$
    \ENDFOR
    \IF{completed patrol for $g_n^{\tau}$}
      \STATE $h_n^{\tau} \leftarrow [0,\ t]$
      \STATE select the new $g_n^{\tau}$
    \ENDIF
    \IF{$A_n(t)\neq\emptyset$}
      \STATE $H_n^s(t) \leftarrow h_n(t)$ with the $s$ largest $t_n^k(t)$ values
      \STATE send $H_n^s(t)$ to all $r_m\in A_n(t)$
    \ENDIF
    \IF{received $H_m^s(t-1)$ from $r_m \in A_n(t-1)$}
      \FORALL{$h_m^k(t-1) \in H_m^s(t-1)$}
        \IF{$t^k_m(t-1) > t^k_n(t)$}
          \STATE $h^k_n(t) \leftarrow h^k_m(t-1)$
        \ENDIF
      \ENDFOR
    \ENDIF
    \STATE \textbf{return} $H_n(t)$
\end{algorithmic}
\end{algorithm}

The parameter $s$ is determined by the communication bandwidth constraints. If bandwidth is ample, the robots can exchange all items in $H_n(t)$, i.e., $s \geq K$. However, in real-world missions, bandwidth limitations often degrade mission performance or scalability to the map size: $K$. The procedure may restrict $s$ to be less than $K$ to incorporate this viewpoint. In such cases, $r_n$ sends $h_n(t) \in H_n^s(t)$ with the $s$ largest $t_n^k(t)$ values.

Through Algorithm~\ref{alg_udassume}, assumed idleness integrates patrol achievements from multiple robots into a constant-size format. Regardless of the swarm size $N$, each robot's assumed idleness holds $2K$ items, comprising $K$ assumed idleness values and their update time. This approach facilitates the propagation of achievements among all robots, including the BS, with a lower risk of communication congestion. A robot can incorporate other robots' achievements through lines 11-14. The robot subsequently transfers the achievement to another robot through lines 8-10 at the next timestep. Further, the update time of assumed idleness by the BS ($r_1$), $t_1^k(t)$, serves as the latest update time for information about $g^k$. The BS accordingly calculates grid SA delay by $D_{\mathrm{SA}}^k(t) = t - t_1^k(t)$.

\subsection{Patrol Target Grid Selection}
As a main part of LR-PT, each robot $r_n$ independently determines its patrol target grid $g_n^{\tau}$. Unlike systems reliant on centralized control, each robot bases its decision on its local knowledge $H_n(t)$. $r_n$ also considers another status parameter: report priority $p_n(t)$, which reflects the robot's perceived urgency to report its patrol achievements to the BS. A higher $p_n(t)$ increases the likelihood that $r_n$ will select $g_n^{\tau}$ closer to the BS. Patrolling the area closer to the BS leads to reporting robot's achievement to the BS.

\subsubsection{Report Priority}
The concept of report priority $p_n(t)$ encapsulates each robot's obligation to communicate with the BS to report mission progress. This procedure also facilitates distributed area partitioning and allocation within LR-PT. $p_n(t)$ works as positive feedback, encouraging robots that have recently communicated with the BS to do so again. Conversely, it deters the robots from nearing the BS if they have not recently communicated with the BS. This procedure leads to an emergent area partitioning or role allocation: some robots assume the role of \textit{explorers}, focusing their patrol on distant areas from the BS. Other robots act as \textit{reporters}, patrolling closer regions and relaying the achievements of explorers to the BS. When an explorer and a reporter connect, they exchange information via Algorithm~\ref{alg_udassume}, and the reporters owe the explorers' duty to report patrol achievements for distant regions.

Algorithm~\ref{alg_LRPT} elaborates on the maintenance of $p_n(t)$. Briefly, $r_n$ increments $p_n(t)$ over time, subject to a maximum value: $p_M$ (lines 2-3). The procedure corresponds to the fact that the need to report to the BS grows according to the time elapses. $p_M$ is the parameter that determines how early the robot saturates its report priority, whereas the smaller $p_M$ promotes the reporting behaviors. $p_M$ was optimized with consideration for balance with other parameters before operations. Upon receiving a message from another robot $r_m \in A_n(t-1)$, $r_n$ checks the messages from $r_m$ (lines 4-13). If $r_m$ is the BS, $r_n$ updates its last contact timestamp: $\omega_n^b(t)$ (lines 7-8). If not, $r_n$ compares $\omega_n^b(t)$ with $\omega_m^b(t-1)$, adopting $r_m$'s report priority if $r_n$'s last BS contact was more recent (lines 9-11). This process corresponds to the procedure that $r_n$ has taken on $r_m$'s reporting duty. $\eta$ is the discount rate for this process. The larger $\eta$ promotes reporter's behavior that gets closer to the BS upon connections to an explorer, and is optimized with consideration for balance with other parameters. If no exchange occurs with any robots within $A_n(t-1)$, $r_n$ resets $p_n(t)$ to zero, assuming another robot has owed its report priority (lines 12-13). Symmetrically, $r_n$ broadcasts its $p_n(t)$ and $\omega_n^b(t)$ to $\forall r_m \in A_n(t)$ (line 14).

\begin{algorithm}[htbp]
\caption{Algorithm to update $p_n(t)$ according to LR-PT at time $t$}
\label{alg_LRPT}
\begin{algorithmic}[1]
  \IF{$r_n$ is NOT connected to the BS}
    \STATE $p_n(t) \leftarrow \mathrm{min}\{p_M,\ p_n(t-1)+1\}$
  \ENDIF
  \IF{$A_n(t-1) \neq \emptyset$}
    \STATE is\_update $\leftarrow$ false
    \FORALL{$r_m \in A_n(t-1)$}
      \IF{$r_m$ is BS}  
        \STATE $\omega_n^b \leftarrow t$
      \ELSIF{$\omega_m^b(t-1) < \omega_n^b(t)$}
        \STATE $p_n(t) \leftarrow \mathrm{max}\{p_n(t),\ p_m(t-1)\} + \eta\mathrm{min}\{p_n(t),\ p_m(t-1)\}$
        \STATE is\_update $\leftarrow$ true
      \ENDIF
    \ENDFOR
    \IF{is\_update $==$ false}
      \STATE $p_n(t) \leftarrow 0$
    \ENDIF
  \ENDIF
  \STATE send $p_n(t)$ and $\omega_n^b(t)$ to $\forall r_m \in A_n(t)$
  \STATE \textbf{return} $p_n(t)$
\end{algorithmic}
\end{algorithm}

By Algorithm~\ref{alg_LRPT}, when a robot $r_n$, having recently communicated with the BS, encounters another robot $r_m$, it incorporates $p_m(t-1)$ into $p_n(t)$. This enhanced $p_n(t)$ motivates $r_n$ to communicate with the BS again by influencing its patrol target selection. The mechanism acts as positive feedback, encouraging $r_n$ to serve both as a reporter to the BS and a patroller of the proximal areas to the BS, and vice versa for $r_m$ as an explorer.
\\
\subsubsection{Evaluation of Patrol Target Grid Candidates}
When selecting patrol target grids, $r_n$ selects the grid that maximizes utility, as given by:
\begin{equation}
    g_n^{\tau} = \underset{g^k \in G_n^{\delta}(t)} {\operatorname{argmax}}\ U_n^k(t)
\end{equation}
Here, $G_n^{\delta}(t)$ represents the set of grids within a distance $\delta$ from $\textit{\textbf{x}}_n(t)$. It corresponds to the consideration range for each robot upon patrol target grid selection, which affects the computational costs and patrol comprehensiveness of the algorithm. In current settings, it is set as $\delta=d_c$, reflecting the fact that a robot may acquire information within the distance $d_c$ from other robots via communications. The utility of a grid $g^k$ for $r_n$ at time $t$, denoted $U_n^k(t)$, incorporates the robot's assumed idleness and report priority:
\begin{subequations} \label{eq_LRutility}
  \begin{equation}
    U_n^k = \alpha_n^k(t) \frac{i_n^k(t) + \Delta _n^k(t)}{\Delta_n^k(t)}
  \end{equation}
  \begin{equation}
    \alpha_n^k(t) = \mathrm{exp}\left\{-\frac{(\mathrm{max}\{x^k, y^k\} - (p_M-p_n(t)))^2}{2\sigma^2}\right\}
  \end{equation}
\end{subequations}
In the first equation, $r_n$ evaluates $g^k$ by its assumed idleness $i_n^k(t)$ and the expected travel time $\Delta_n^k(t)$ to the grid. The adjustment factor: $\alpha_n^k(t)$ incorporates the report priority: $p_n(t)$ into the utility via the second equation. 

\begin{figure}[ht]
\centering
  \begin{minipage}[b]{0.47\linewidth}
    \centering
    \includegraphics[keepaspectratio, scale=0.55]{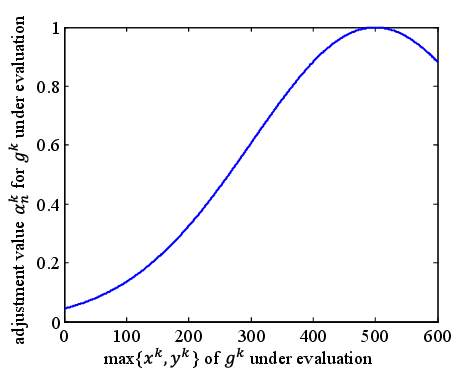}
    \caption{An example plot of $\alpha_n^k(t)$ when $p_n(t) = 100$. $p_n(t)$ is small (i.e., $r_n$ is less required to contact the BS), and it rates $\alpha_n^k(t)$ for the grid with larger x or y coordinate higher.}
    \label{fig_examplealpha_a}
  \end{minipage}
  \hspace{0.03\columnwidth}
  \begin{minipage}[b]{0.47\linewidth}
    \centering
    \includegraphics[keepaspectratio, scale=0.55]{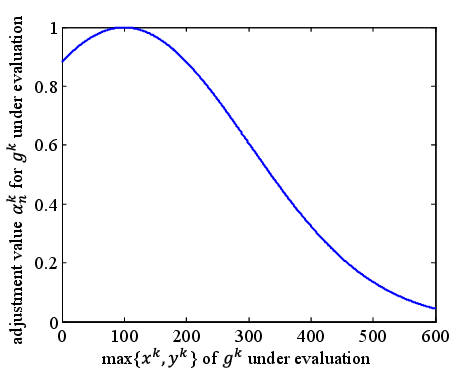}
    \caption{A plot of $\alpha_n^k(t)$ at $p_n(t) = 500$. $p_n(t)$ is large (i.e., $r_n$ is required to contact the BS), and it rates $\alpha_n^k(t)$ for the grid with smaller x or y coordinates higher.}
    \label{fig_examplealpha_b}
  \end{minipage}
\end{figure}

The calculation of $\alpha_n^k(t)$ considers the Chebyshev distance $\mathrm{max}\{x^k, y^k\}$ between the origin and $g^k$. It enhances the utility of grids closer to the BS when $p_n(t)$ is high, indicative of a strong requirement to report its achievements. Conversely, when with a lower $p_n(t)$, indicating a reduced need for immediate communication with the BS, the utility favors grids located farther from the BS. We adopted Chebyshev distance instead of Euclidean distance, to avoid robots' overcrowding to the narrow space around the top-right corner. The parameter $\sigma$ adjusts the sensitivity of $\alpha_n^k(t)$ to the grid's location and is optimized with consideration for balance with other parameters. Figure~\ref{fig_examplealpha_a} and \ref{fig_examplealpha_b} illustrate two example plots of $\alpha_n^k(t)$ with $p_M=600,\ \sigma=200$. In Figure~\ref{fig_examplealpha_a}, a lower $p_n(t)$ results in higher utility ratings for grids located farther from the BS. Conversely, Figure~\ref{fig_examplealpha_b} shows that when $p_n(t)$ is high, grids closer to the BS are rated more favorably. $r_n$ finally selects an adjacent grid that is on the way to the selected grid based on the utility function, as its temporary target. Upon completing a patrol visit to the temporary target grid, the robot recalculates the utilities of neighboring grids again. 

\section{SIMULATIONS}\label{Sim}
The efficacy of LR-PT was assessed via computer simulations replicating patrol missions. This section outlines the configurations and results of the simulations. The simulator is implemented in a decentralized manner using MATLAB, treating each robot as an independent object. Robots could access information within their sensor range and process received messages. The following subsections describe (i)configurations, (ii)performance, (iii)robustness, and (iv)a discussion of the LR-PT's limitations.

\subsection{Configurations}
The following subsection describes the configurations during the simulations. The parameter values define the mission settings and adjust algorithm to the settings. This research introduces several existing algorithms to compare patrol performance to LR-PT.
\\
\subsubsection{Simulation Parameters}
Tables \ref{tab1} and \ref{tab2} present the configurations for the robots and the missions. The mission environment is a square map composed of 400 grids, each grid measuring 30[m] by 30[m]. The sizes of robotic swarms range from $N = 5, 10, 15$, determined to balance computational feasibility, meaningful representation of different scales of robot swarms, and the availability or affordability of acquiring such numbers of robots to survey the designated area. The core parameters for algorithm: $\eta$, $p_M$, and $\sigma$ were decided via pattern search optimization targeting $I_G$ and $I_W$. The normalization of the performance metrics further evaluates whether the algorithm is robust and scalable to swarm size $N$. As for the paramter $s = 8$ as a case with communication bandwidth constraint, its effect is reviewd in Section 5.3.

\begin{table}[t]
	\begin{center}
		\caption{Robot configurations.}
		\label{tab1}
		\renewcommand{\arraystretch}{1.5}
        \small
		\begin{tabular}{|c|c|c|l|}
			\hline
			\textbf{Symbol} & \textbf{Value} & \textbf{Unit}& 
			\textbf{Comment}    \\ \hline
			$v_M$ & 1.5 & \textit{m/sec} & Max linear velocity \\ \hline
			$\phi_M$ & 1.0 & \textit{rad/sec} & Max angular velocity \\ \hline
			$d_c$ & 180 & \textit{m} & communication range \\ \hline
			$d_s$ & 90 & \textit{m} & sensor range \\ \hline
			$\delta$ & 180 & \textit{m} & search range for grids \\ \hline
			$\eta$ & 0.55, 0.40, 0.50 & - & discount rate of $p_n(t)$ \\
                   &                  &   & for $N=5,\ 10,\ 15$       \\ \hline
			$p_M$ & 1088, 703, 425 & - & Max value of $p_n(t)$ \\
                  &                &   & for $N=5,\ 10,\ 15$   \\ \hline
			$\sigma$ & 356, 304, 525 & - & gradient factor of $p_n(t)$ \\
                     &                  &   & for $N=5,\ 10,\ 15$       \\ \hline
			$s$ & $K,\ 8$ & \textit{item/s} & communication \\
			    &         &                 & bandwidth constraint \\ \hline
		\end{tabular}
	\end{center}
\end{table}

\begin{table}[t]
	\begin{center}
		\caption{Mission configurations.}
		\label{tab2}
		\renewcommand{\arraystretch}{1.5}
        \small
		\begin{tabular}{|c|c|c|l|}
			\hline
			\textbf{Symbol} & \textbf{Value} & \textbf{Unit}& 
			\textbf{Comment}    \\ \hline
			$N$ & 5,10,15 & robots & includes the BS \\ \hline
			$K$ & $20 \times 20$ & grids & - \\ \hline
			-   & 30 & \textit{m} & grid size \\ \hline
			$\rho$ & 3 & \textit{m} & patrol completion\\
			       &   &            & threshold \\ \hline
			$T$ & 43200 & \textit{sec} & mission duration \\ \hline
			-   & 60 & trials & number of trials \\ \hline
		\end{tabular}
	\end{center}
\end{table}

At the beginning of a mission ($t_0 = 0$), the simulator locates $r_1$ as the BS at the origin. The remaining $N-1$ robots are randomly located within a fan-shaped area: $\|\textit{\textbf{x}}_n(0)\| \leq 2\sqrt{N}$. All idleness values $i^k(0)$ are set to zero. At this moment, the robots $r_{2}, \dots ,r_{N}$ select their initial patrol target grids and proceed with their patrolling task according to LR-PT. The evaluation consisted of 30 trials, spanning ten trials across three different swarm sizes.
\\
\subsubsection{Compared Algorithms}
This study introduces patrol algorithms for comparison which share similar problem settings, assumptions, and performance metrics, to LR-PT. The introduced algorithms are: (i)Expected Reactive (ER)~\cite{Yan2016}, (ii)DTAP~\cite{Farinelli2017}, (iii)Layered Patrol with Continuous Conenctivity (LPCC)~\cite{Kobayashi_2024jrm}. LPCC is a distributed algorithm specifically designed to maintain the superior SA of a BS. Though algorithms other than LPCC do not consider BS's SA, we prioritized the distributed and reactive nature of the algorithm in order to preserve the advantage of swarm robotics. Thus, comparisons with algorithms relying on centralized coordination, preliminary optimization, or assumptions of information availability that are not realistic in practical scenarios, are beyond the scope of this research. 

For the purposes of this comparison, the following modifications were applied to each algorithm to align them more closely with LR-PT’s operational context:
\begin{itemize}
    \item Deploy one robot as a BS at the origin.
    \item The communication range is limited, mirroring the setup in LR-PT.
    \item Each robot maintains assumed idleness and shares it with the BS upon connection, facilitating the computation of $D_{\mathrm{MSA}}$ and $D_{\mathrm{WSA}}$.
\end{itemize}
Other relevant parameters were determined using a pattern search optimization to ensure each compared algorithm performs optimally in terms of graph idleness and worst idleness.

\subsection{Performance Evaluation}
This subsection describes the simulation results and discusses the performance of LR-PT. The first part visualizes the robots' behavior, while the following parts describe the mission performance based on multiple metrics.
\\
\subsubsection{Simulation Overview}
\begin{figure}[t]
    \centering
    \resizebox*{8cm}{!}{\includegraphics{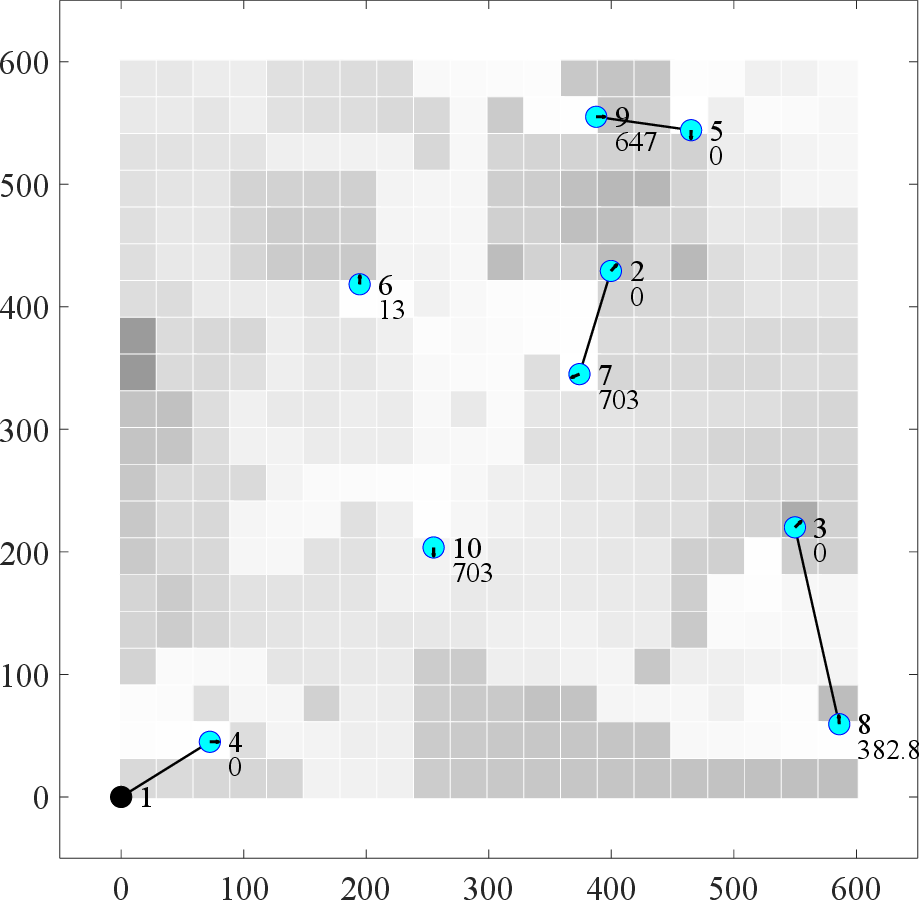}}
    \caption{A screenshot of the simulated patrol mission with $N=10$. The black circle at the bottom left is the BS, and the cyan circles represent patrolling robots. The color of the grids indicates their idleness level; darker colors signify higher idleness.}
    \label{fig_scshot}
\end{figure}

LR-PT successfully conducted the patrol missions. Figure~\ref{fig_scshot} captures a moment from a trial involving a swarm of $N=10$: one BS ($r_1$, shown as a black circle), and nine patrolling robots ($r_{2}, \dots ,r_{10}$, shown as the cyan circles). The lines denote the network connections. The numbers appended to each robot show its ID in the upper row and report priority: $p_n(t)$ in the lower row. The grid colors reflect their idleness ($i^k(t)$), with darker colors indicating higher idleness. It is worth to note that each robot do not necessarily assumes the values of the grids accurately as shown in the figure, since they can accurately realize the value only when they arrive at the grids.

In the figure, $r_4$ is directly connected to the BS, sharing its assumed idleness. Robots $r_2$ and $r_7$ are interconnected at this moment. Although not visible in the figure, $r_7$ received $r_2$'s latest contact time to the BS $\omega_2^b(t-1)$ and report priority $p_2(t-1)$ (information at timestep: $t-1$ is available for $r_7$, since the message transmissions require a single timestep per hop in the current problem setting). $r_7$ realized that $\omega_7^b(t)$ is larger than $\omega_2^b(t-1)$: $r_7$ has connected to the BS more recently than $r_2$. In such case, $r_7$ incorporates $p_2(t-1)$ into $p_7(t)$ according to the lines 9-10 in Algorithm~\ref{alg_LRPT}. Reversely, $r_2$ entrusted its report priority to $r_7$ and refresh $p_2(t)$ as zero.
\\
\subsubsection{Idleness Performance}
\begin{figure}[t]
  \centering
  \resizebox*{8cm}{!}{\includegraphics{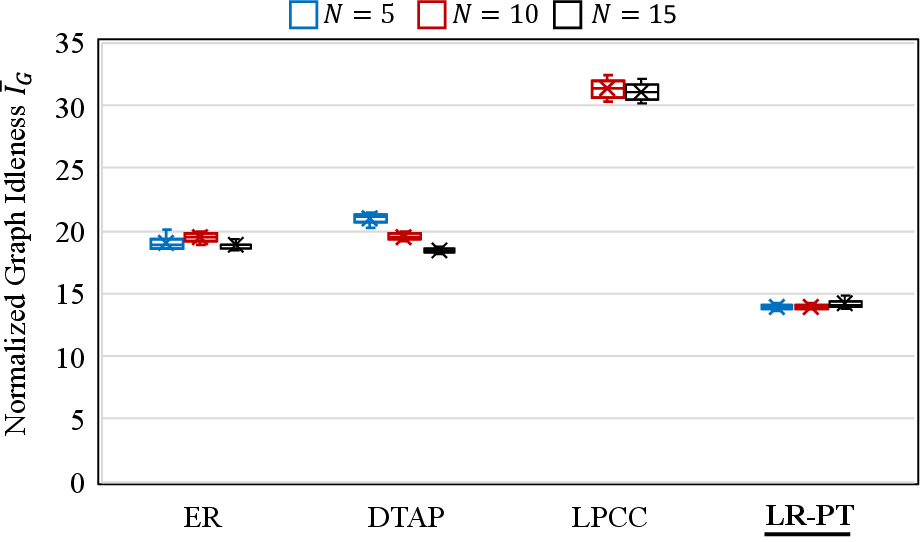}}
  \caption{Performance from an normalized graph idleness perspective, across various swarm sizes $N$. The smaller values indicate the better performance. No data for LPCC with $N=5$. The bold text indicates the proposed method.}
  \label{fig_gidle_performance}
\end{figure}

\begin{figure}[t]
  \centering
  \resizebox*{8cm}{!}{\includegraphics{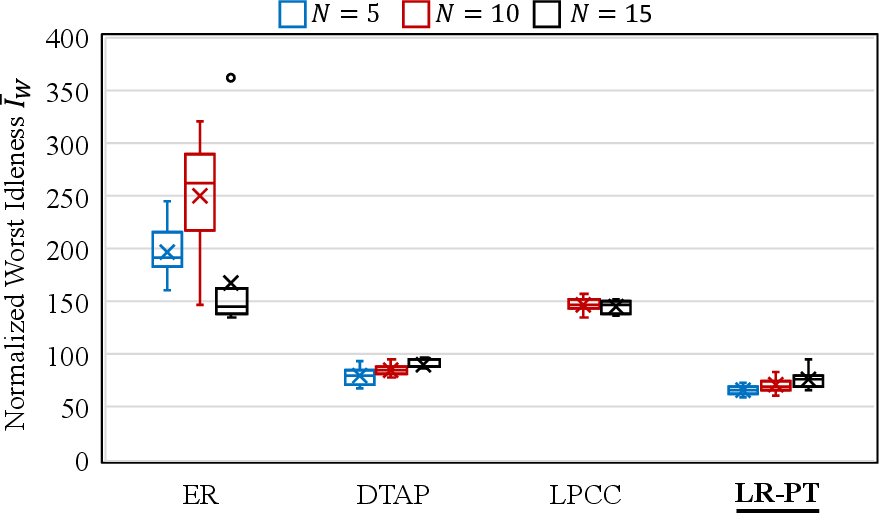}}
  \caption{Performance from an normalized worst idleness perspective, across various swarm sizes $N$. The smaller values indicate the better performance. No data for LPCC with $N=5$. The bold text indicates the proposed method.}
  \label{fig_widle_performance}
\end{figure}

Figures~\ref{fig_gidle_performance} and \ref{fig_widle_performance} present the patrol performance in terms of normalized graph idleness: $\overline{I}_G$ and normalized worst idleness: $\overline{I}_W$. Due to its connectivity constraint, LPCC failed to patrol the distant area from the BS with the swarm size $N=5$, and thus, the data is not shown in the figures. To ensure that the metrics reflect the algorithm's intrinsic performance, initial data of each trial, representing the transitional phase, was excluded. The simulations omitted such phase by setting $t_0 = 10000$ when computing the metrics, focusing on steady-state performance.

LR-PT exhibited superior performance across both metrics when compared to the other algorithms. Figures~\ref{fig_gidle_performance} and Figure~\ref{fig_widle_performance} display the performance for each algorithm across different swarm sizes. Notably, LR-PT outperformed the second-best algorithm, DTAP, by $20-30\%$ in terms of normalized graph idleness: $\overline{I}_G$. The algorithm also performed better by $15-20\%$ in normalized worst idleness: $\overline{I}_W$. Furthermore, LR-PT significantly outperformed LPCC, another distributed algorithm that considers BS's SA. The proposed algorithm almost halved the metric values of LPCC, indicating that the robots visited each location twice as often. The better performance of the proposed LR-PT can be attributed to the design of the utility function, incorporating both the assumed idleness and travel cost. Though ER and DTAP also performed well, the update of assumed idleness at every timestep shown in Algorithm~\ref{alg_udassume} also contributed to LR-PT's better performance. 

LR-PT also demonstrated an almost scalable performance for the swarm size $N$. From its definition, the constant value of the normalized metrics indicates that the swarm performed proportionately to its sizes, which is called \textit{scalable} in this context. With this characteristic, deploying more robots leads to a commensurate improvement in performance. LR-PT performed with scalability in terms of $\overline{I}_G$. Since the assumed idleness held by each robot can incorporate other robots' patrol achievements via communication, the robots can avoid duplicate patrolling and thus maintain scalability to the swarm sizes. On the other hand, the swarm with $N=15$ under LR-PT have performed slightly worse than the swarm with $N=5$, in terms of $\overline{I}_W$. The locality of robots' behavior controlled by the consideration of travel cost and the parameter $\delta$ may led to this trend.
\\
\subsubsection{SA performance}
\begin{figure}[t]
  \centering
  \resizebox*{8cm}{!}{\includegraphics{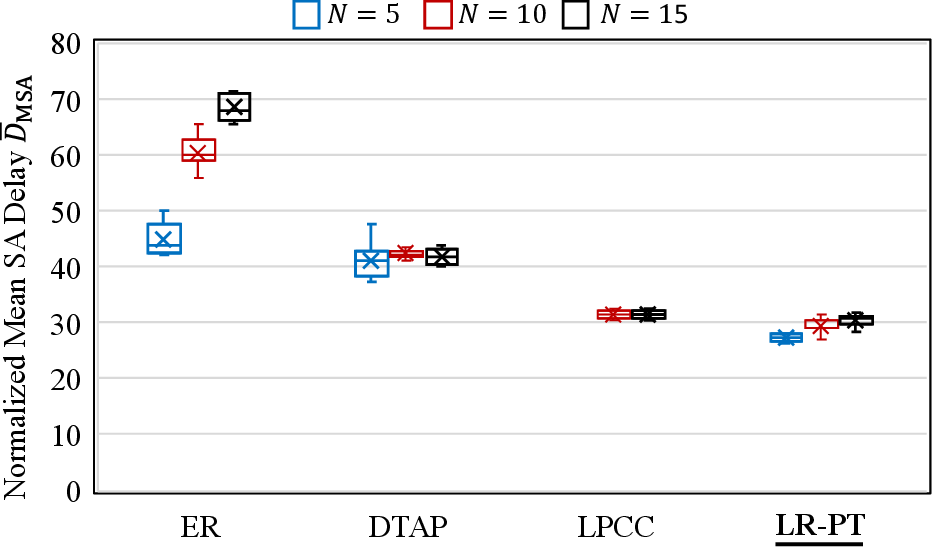}}
  \caption{Performance from an Mean SA delay perspective, across various swarm sizes $N$. The smaller values indicate the better performance. No data for LPCC with $N=5$. The bold text indicates the proposed method.}
  \label{fig_msa_performance}
\end{figure}

\begin{figure}[t]
  \centering
  \resizebox*{8cm}{!}{\includegraphics{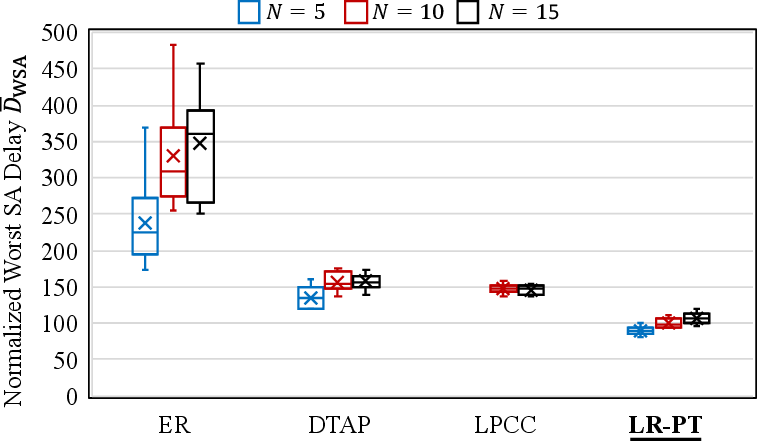}}
  \caption{Performance from an Worst SA delay perspective, across various swarm sizes $N$. The smaller values indicate the better performance. No data for LPCC with $N=5$. The bold text indicates the proposed method.}
  \label{fig_wsa_performance}
\end{figure}

LR-PT also exhibited superior performance in terms of the SA metrics. Figures~\ref{fig_msa_performance} and \ref{fig_wsa_performance} show the algorithms' performance in terms of normalized mean SA delay: $\overline{D}_{\mathrm{MSA}}$ and normalized worst SA delay: $\overline{D}_{\mathrm{WSA}}$. Though LR-PT are free of connectivity constraint, the algorithm have drived the robots to efficiently report their patrol achievements to the BS. The performance was about $5\%$ and $30\%$ better than the second best: LPCC in terms of $\overline{D}_{\mathrm{MSA}}$ and $\overline{D}_{\mathrm{WSA}}$, respectively. Due to the fact that ER and DTAP do not consider BS's SA, they performed worse than LR-PT in this aspect. Again, this research prioritize the comparison among the distributed algorithm to focus on both evaluating patrol performance and preserving advantages of swarm robotics.

LR-PT also demonstrated an almost scalable performance for the swarm size $N$. The almost constant values demonstrated in Figures \ref{fig_msa_performance} and \ref{fig_wsa_performance} indicate that deploying more robots is worthwhile in terms of these performance metrics. The slight increase in the metrics can be attributed to the computation method of SA delay, which consolidates multiple visits to a grid to a single item of assumed idleness. For instance, if the robots visit a grid $g^k$ at times $t_1$ and $t_2 (> t_1)$ without reporting these visits to the BS, the grid idleness $i^k(t)$ is reset to zero at each visit, contributing to a reduction in $I_G$ and $I_W$. On the other hand, the recorded assumed idleness for the grid, $h_n^k(t)$, only reflects the most recent visit at $t_2$ ($t_n^k(t) = t_2$). Upon connecting to the BS ($r_1$), only the latest timestamp ($t^k_1(t) = t_2$) is used to compute the grid's SA delay. Consequently, the BS can no longer be aware of the reduction in $i^k(t)$ from $t_1$ to $t_2$, and the SA delay tends to be larger. This trend becomes more significant with larger swarms, meaning that the frequency of visits for a specific grid surpasses the frequency of reports about that grid as the swarm size increases.
\\
\subsubsection{Emergent Area Partitioning}
\begin{figure}[t]
  \centering
  \resizebox*{9cm}{!}{\includegraphics{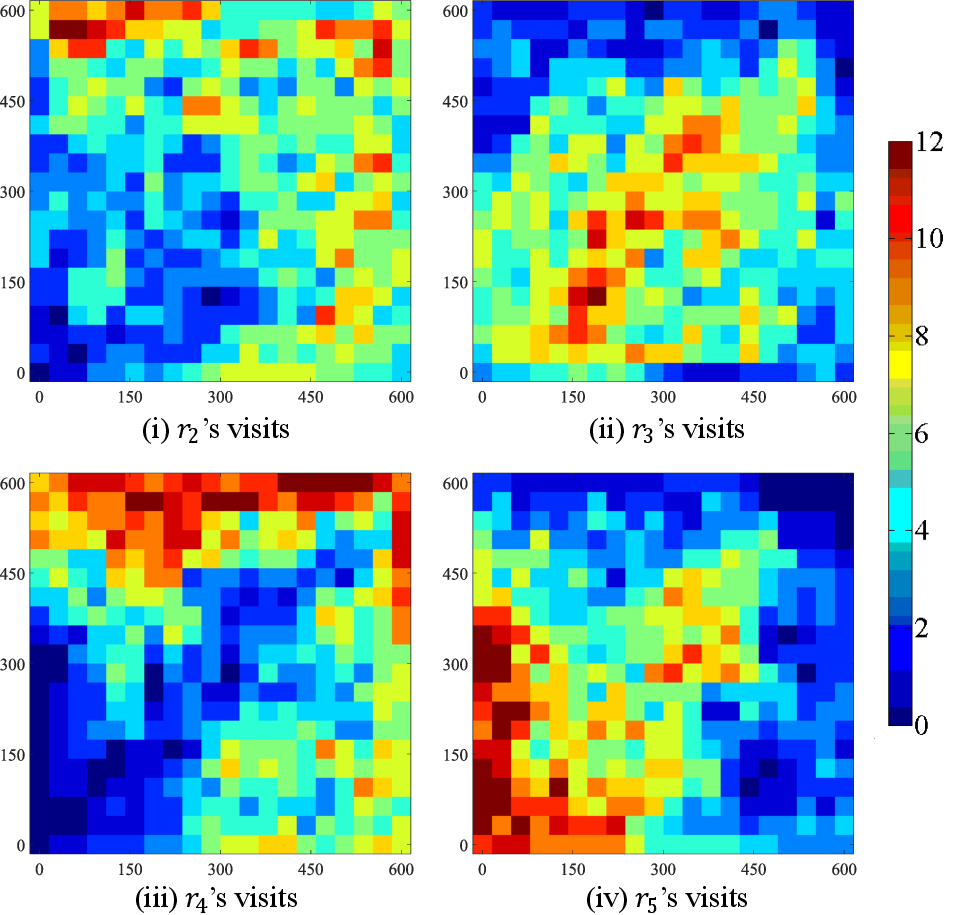}}
  \caption{Heatmaps for the number of patrols for each grid by each robot except for the BS during a single trial of mission with $N=5$.}
  \label{fig_areapartitionN5}
\end{figure}

\begin{figure}[t]
  \centering
  \resizebox*{8cm}{!}{\includegraphics{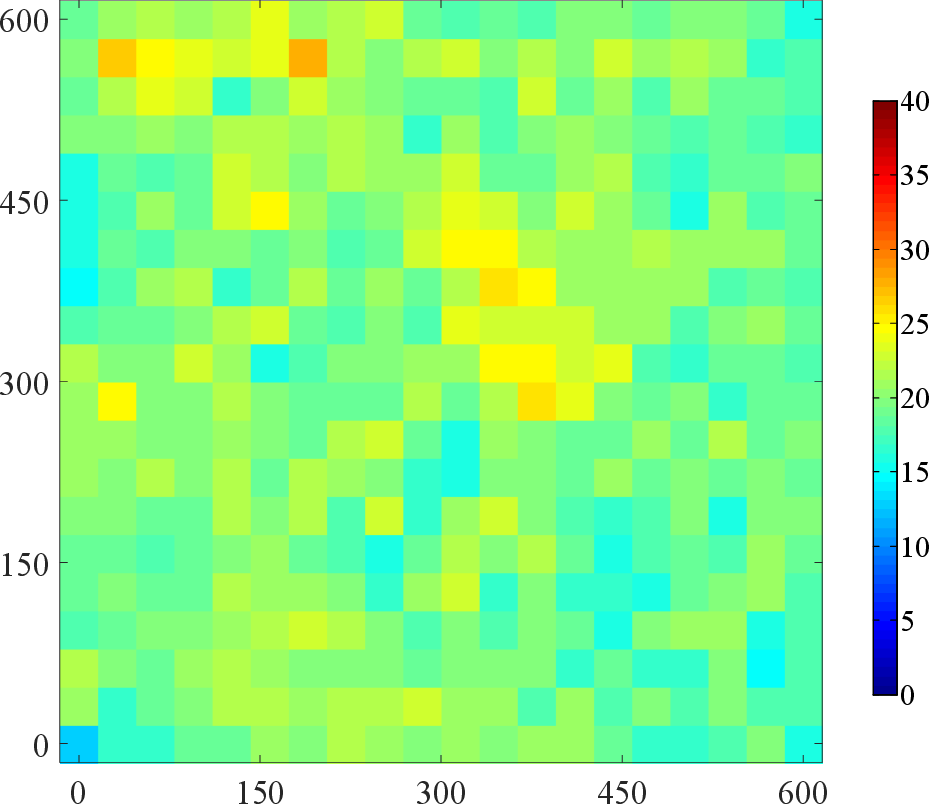}}
  \caption{A heatmap for the total number of patrols for each grid by all the robots during a single trial of mission with $N=5$.}
  \label{fig_areapartitionN5total}
\end{figure}

Autonomous area partitioning emerged as an additional characteristic of LR-PT. Figures~\ref{fig_areapartitionN5} and \ref{fig_areapartitionN5total} display heatmaps of the number of patrols throughout the entire area during a single trial, as a typical case. In Figure~\ref{fig_areapartitionN5}, $r_3$ and $r_5$ acted as reporters, patrolling areas close to the BS. $r_2$ and $r_4$ covered areas far from the BS as explorers, delegating their reporting duties to the reporters. When they encounter, explorers' report priorities are transferred to reporters, and thus, the robots keep their current roles. Despite this partitioning, the total number of patrols by all the robots did not exhibit significant bias across the map, as shown in Figure~\ref{fig_areapartitionN5total}.

\begin{figure}[t]
  \centering
  \resizebox*{9cm}{!}{\includegraphics{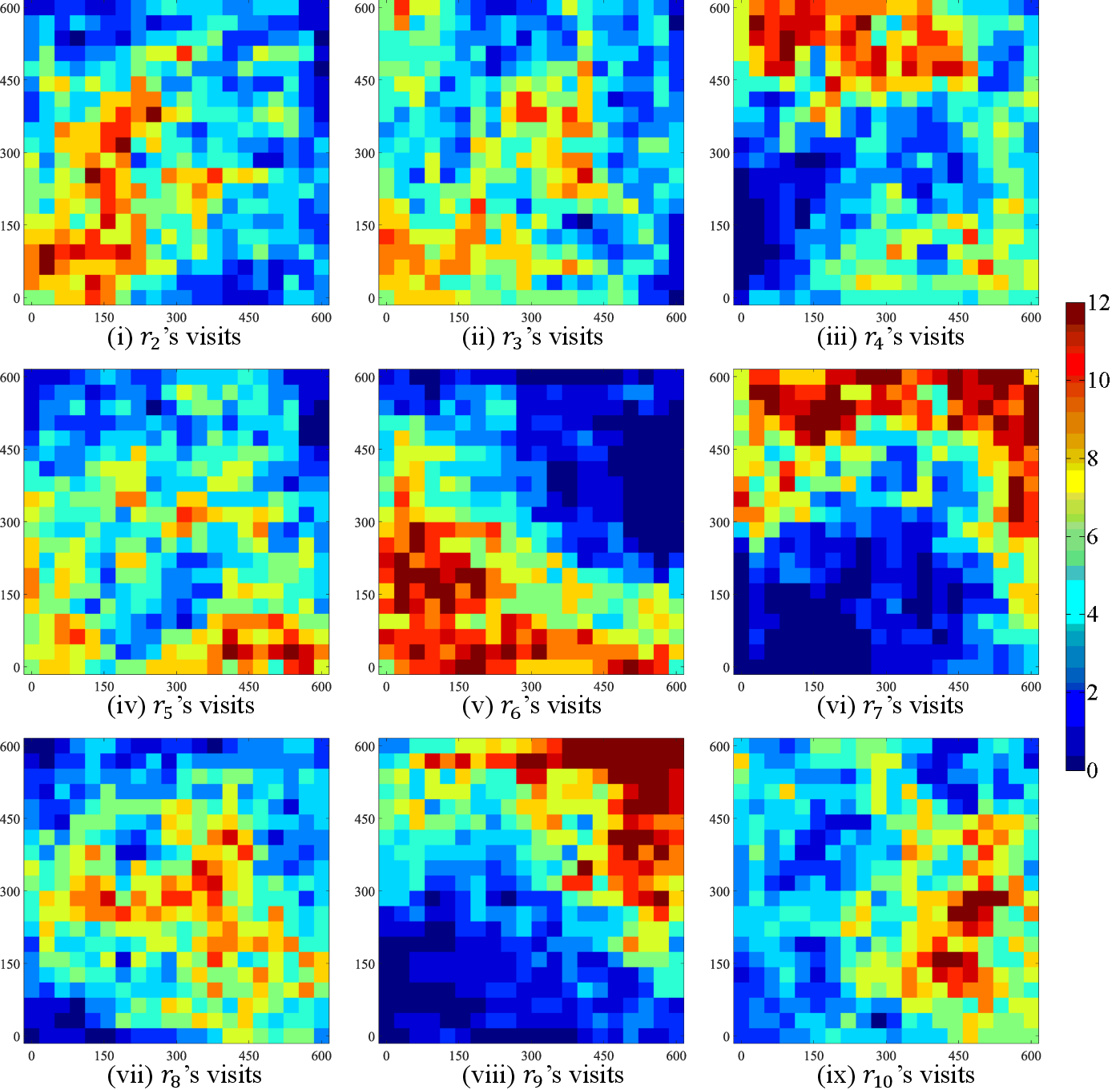}}
  \caption{Heatmaps for the number of patrols for each grid by each robot except for the BS during a single trial of mission with $N=10$.}
  \label{fig_areapartitionN10}
\end{figure}

\begin{figure}[t]
  \centering
  \resizebox*{8cm}{!}{\includegraphics{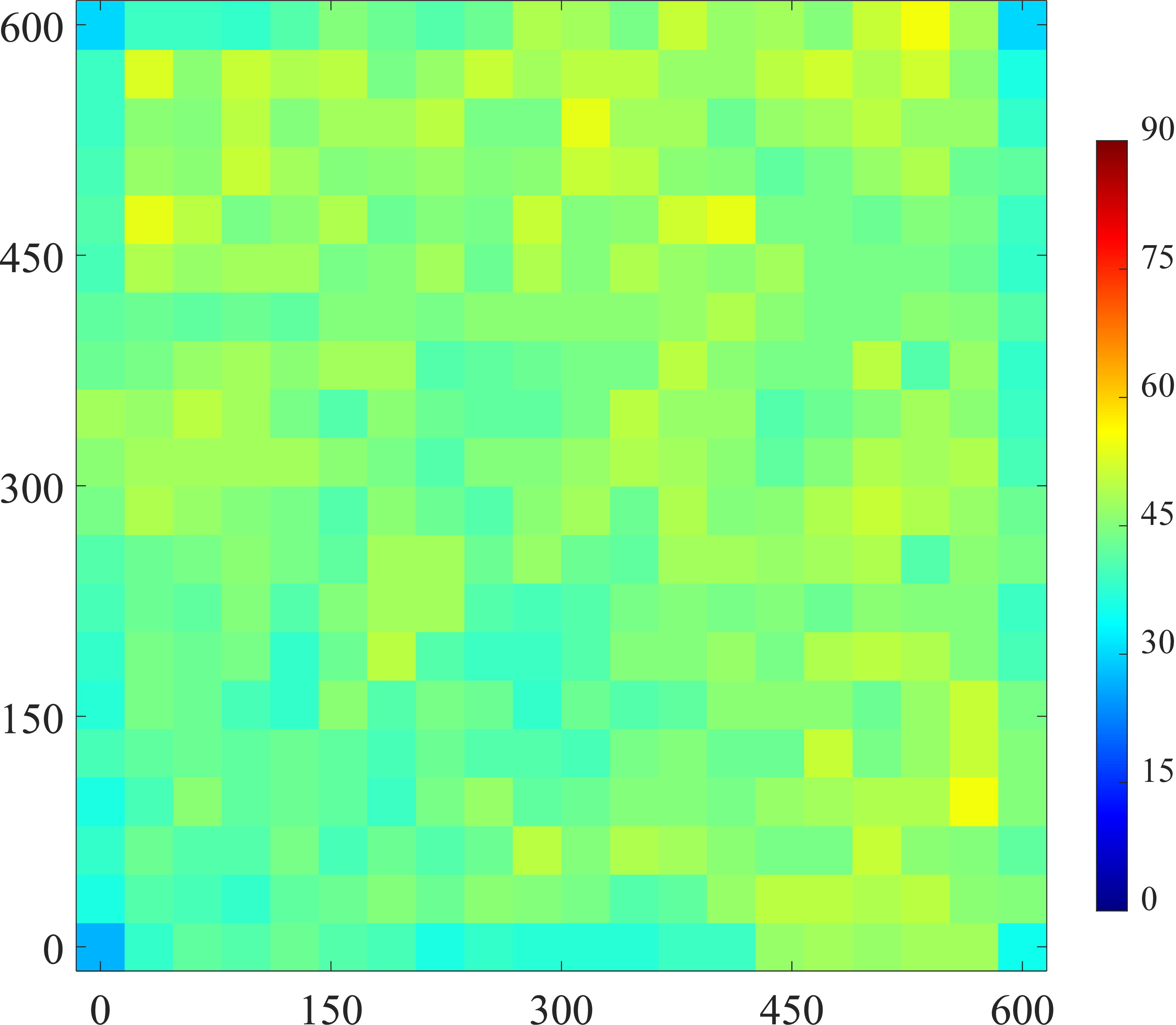}}
  \caption{A heatmap for the total number of patrols for each grid by all the robots during a single trial of mission with $N=10$.}
  \label{fig_areapartitionN10total}
\end{figure}

\begin{figure}[t]
  \centering
  \resizebox*{9cm}{!}{\includegraphics{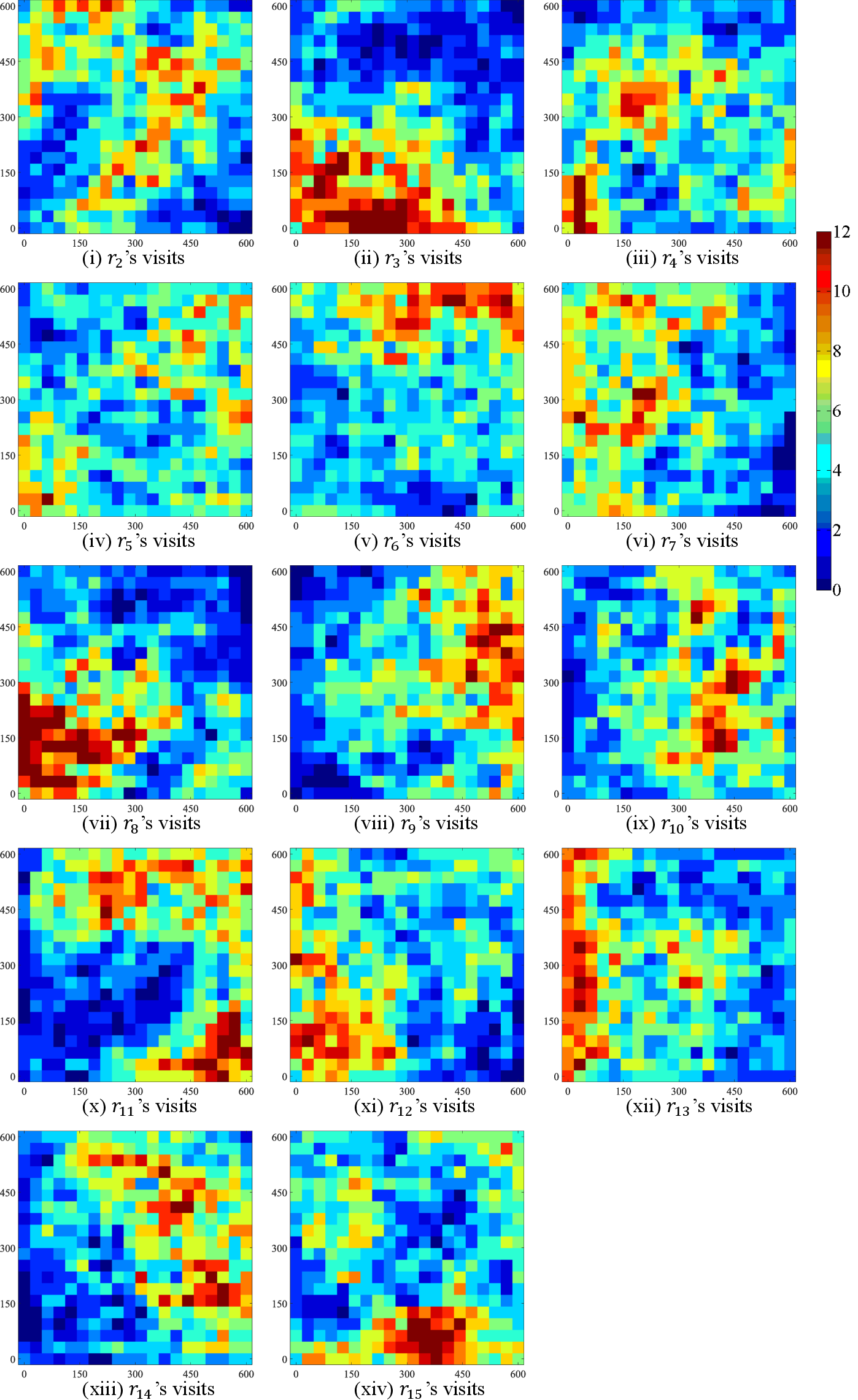}}
  \caption{Heatmaps for the number of patrols for each grid by each robot except for the BS during a single trial of mission with $N=15$.}
  \label{fig_areapartitionN15}
\end{figure}

\begin{figure}[t]
  \centering
  \resizebox*{8cm}{!}{\includegraphics{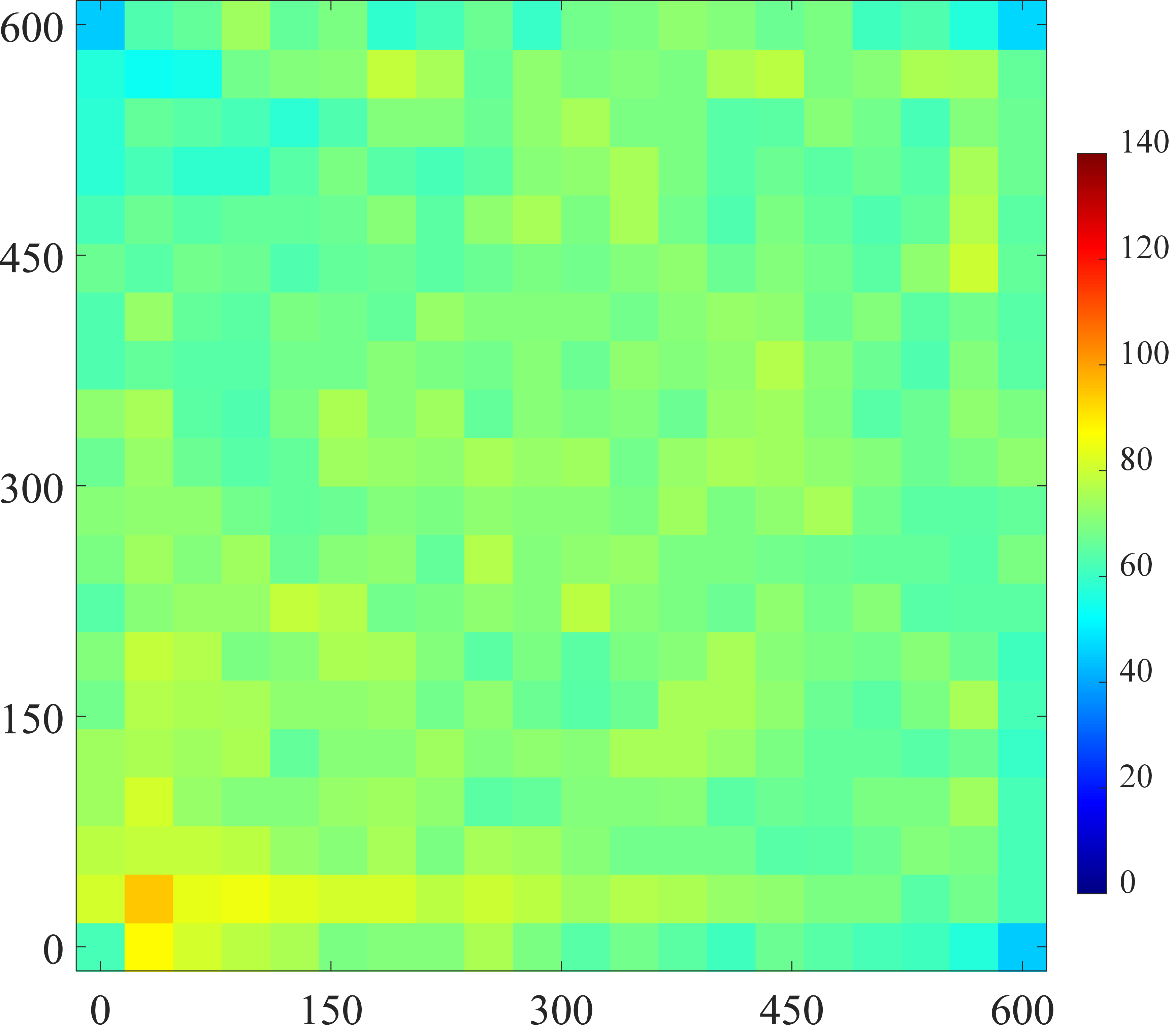}}
  \caption{A heatmap for the total number of patrols for each grid by all the robots during a single trial of mission with $N=15$.}
  \label{fig_areapartitionN15total}
\end{figure}

Figures \ref{fig_areapartitionN10}, \ref{fig_areapartitionN10total}, \ref{fig_areapartitionN15}, and \ref{fig_areapartitionN15total} further illustrate the area partition by the larger swarms. The same trend of area partitioning has emerged in the larger swarm sizes, allocating the role reporter, explorer, or intermediary of the two roles to each robot. It is worth noting that this partitioning emerged merely from the positive feedback triggered by communications between robots. For this reason, some robots can change roles during the mission so that there is no significant bias in the patrol area over the entire mission period. This can be exemplified by the $r_5$'s heatmap in Figure \ref{fig_areapartitionN15}.

This emergent area partitioning enhances the locality of each robot. For the current approach with a smaller $\delta$, each robot computes utilities for a limited number of grids close to its location. It is advantageous since it ensures scalability in computational costs, but it could potentially lead to local optima. In such a case, there is a possibility that all robots will temporarily move away from an area that has not been patrolled for a long time by a distance of more than $\delta$. This partitioning, however, may avoid such local optima by ensuring that at least one robot is assigned to the area and eventually patrols all grids in the area. This characteristic has been demonstrated in Figure~\ref{fig_wsa_performance}, where the proposed LR-PT outperformed the second-best DTAP by $15-20\%$ in $\overline{I}_W$. Though DTAP considers all grids at patrol target selection and accordingly performed well in patrol comprehensiveness, LR-PT still performed better in such a perspective. While LR-PT only considers grids in the range $\delta$ at patrol target selection, the emergent area partitioning successfully maintained the patrol comprehensiveness. Moreover, in future applications involving robotic learning capabilities, area partitioning could facilitate the acquisition of local and temporary specializations for tasks occurring in each area. 

\subsection{Robustness Evaluation}
This section assesses the algorithm's robustness by simulating real-world scenario with communication bandwidth constraint or another scenario where robots might fail due to malfunctions. 
\\
\subsubsection{Configurations}
The configurations for the robots and simulations remain largely consistent with those outlined in Tables~\ref{tab1} and \ref{tab2}. The algorithm limits the number of items that can be exchanged between robots at line 9 in Algorithm~\ref{alg_udassume} to simulate communication bandwidth constraints. This factor can be modified via the parameter $s$, and it is set as $s=8$ in the trials with communication bandwidth constraint. Since the previous setting was $s=K\ (=400)$, the current setting with constraint corresponds to $1/50$ restriction in terms of bandwidth.

To simulate failures, on the other hand, the mission duration $T$ is extended to $T = 129600$. At $t=43200$, 20\% of the robot population, those with the largest IDs (e.g., $r_9$ and $r_{10}$ in a swarm of $N=10$), are programmed to fail, with recovery scheduled for $t=86400$. Each phase of the simulation, before failure, during failure, and after recovery, is allocated 43200 timesteps, which is identical to the previous set of $T$ without failures. Since the change in performance over time is the main focus, the simulations were conducted with a single trial for each condition.
\\
\subsubsection{Robustness to Communication Bandwidth Constraint}
\begin{figure}[t]
  \centering
  \resizebox*{9cm}{!}{\includegraphics{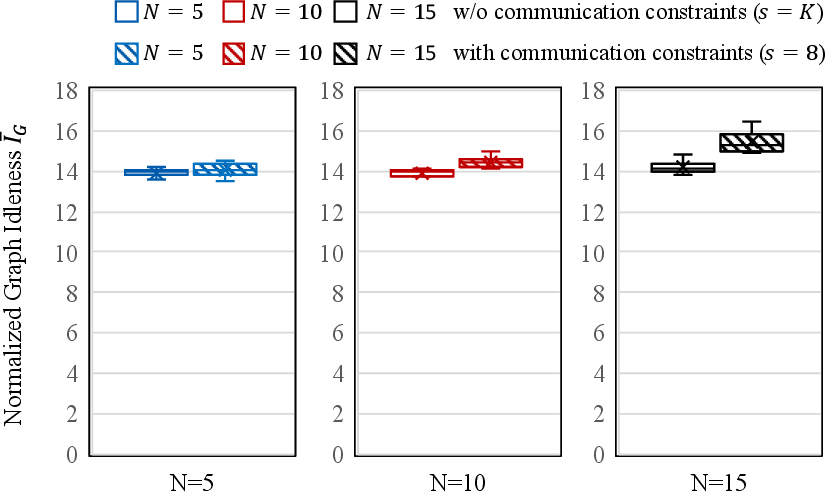}}
  \caption{Performance by LR-PT in graph idleness perspective, with different communication bandwidth constraints ($s$). Constant values highlight the robust performance against the constraints, whereas larger values indicate a degradation in performance.}
  \label{fig_commGI}
\end{figure}

\begin{figure}[t]
  \centering
  \resizebox*{9cm}{!}{\includegraphics{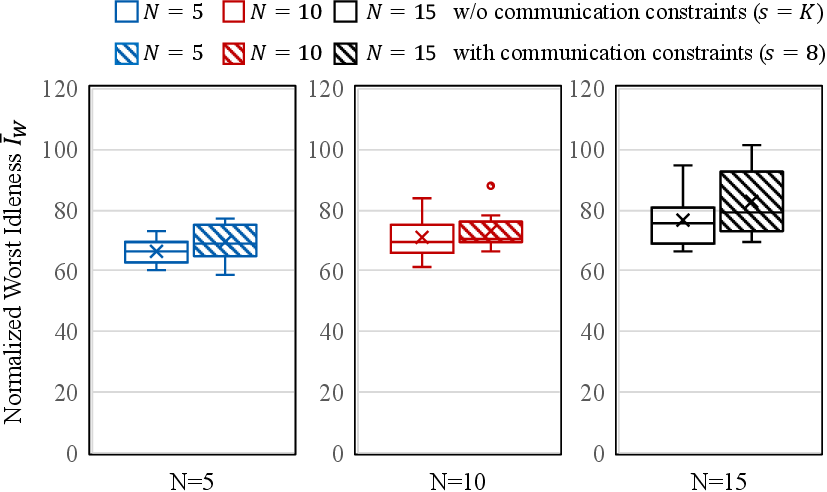}}
  \caption{Performance by LR-PT in worst idleness perspective, with different communication bandwidth constraints ($s$). Constant values highlight the robust performance against the constraints, whereas larger values indicate a degradation in performance.}
  \label{fig_commWI}
\end{figure}

The proposed LR-PT demonstrated robustness to the communication bandwidth constraint. Figures~\ref{fig_commGI} and \ref{fig_commWI} show the idleness-based performance with the different constraint. Though the bandwidth has been restricted as $1/50$ by changing $s=K$ to $s=8$, the performance degradation was only $<10\%$ at worst case, with $N=15$. With the smaller swarms, the idleness-based performance was maintained almost constantly, which indicates the algorithm's robustness to bandwidth constraints. Due to the more dense population in the field, the larger swarms require more sufficient communication between robots to avoid target duplications. This characteristic may have resulted in a relatively worse performance with the larger swarms.

\begin{figure}[t]
  \centering
  \resizebox*{9cm}{!}{\includegraphics{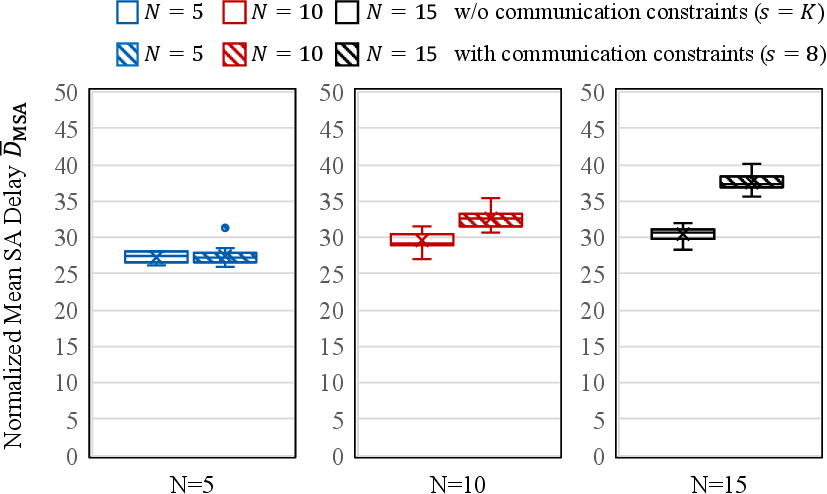}}
  \caption{Performance by LR-PT in Mean SA delay perspective, with different communication bandwidth constraints ($s$). Constant values highlight the robust performance against the constraints, whereas larger values indicate a degradation in performance.}
  \label{fig_commMSA}
\end{figure}

\begin{figure}[t]
  \centering
  \resizebox*{9cm}{!}{\includegraphics{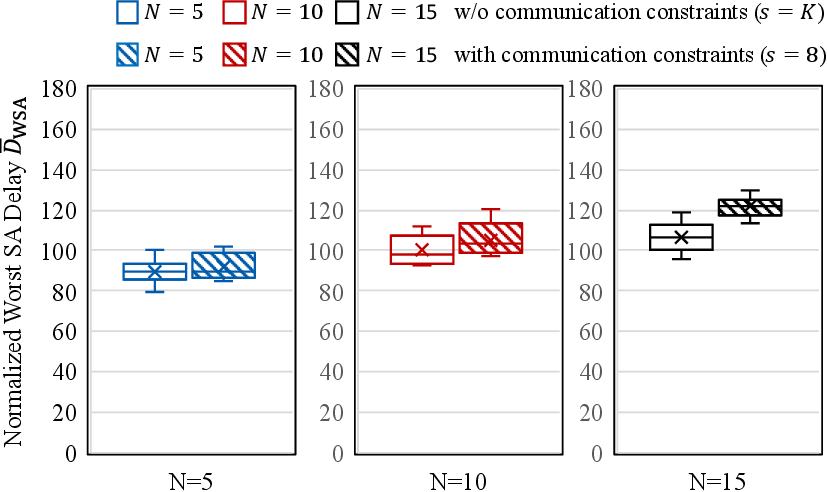}}
  \caption{Performance by LR-PT in Worst SA delay perspective, with different communication bandwidth constraints ($s$). Constant values highlight the robust performance against the constraints, whereas larger values indicate a degradation in performance.}
  \label{fig_commWSA}
\end{figure}

LR-PT also reserved a certain performance even under the communication bandwidth constraint (Figures~\ref{fig_commMSA} and \ref{fig_commWSA}). For larger swarms, which rely more heavily on communication, the reduced bandwidth impacts SA performance, with $\overline{D}_{\mathrm{MSA}}$ and $\overline{D}_{\mathrm{WSA}}$ deteriorating by 15\% and 10\%, respectively, for a swarm size of $N=15$. Modifying the rules for target grid selection to maintain connections until robots update their assumed idleness via communications could improve robustness to communication constraints.
\\
\\

\subsubsection{Robustness to Robot Failures}
\begin{figure}[t]
    \centering
    \resizebox*{8cm}{!}{\includegraphics{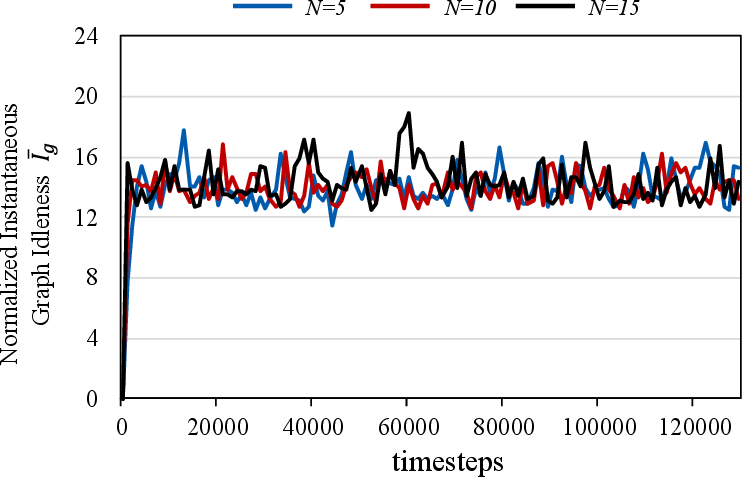}}
    \caption{Changes in normalized instantaneous graph idleness: $\overline{I}_g(t)$ over time, across different swarm sizes.}
    \label{fig_timeIg}
\end{figure}

\begin{figure}[t]
    \centering
    \resizebox*{8cm}{!}{\includegraphics{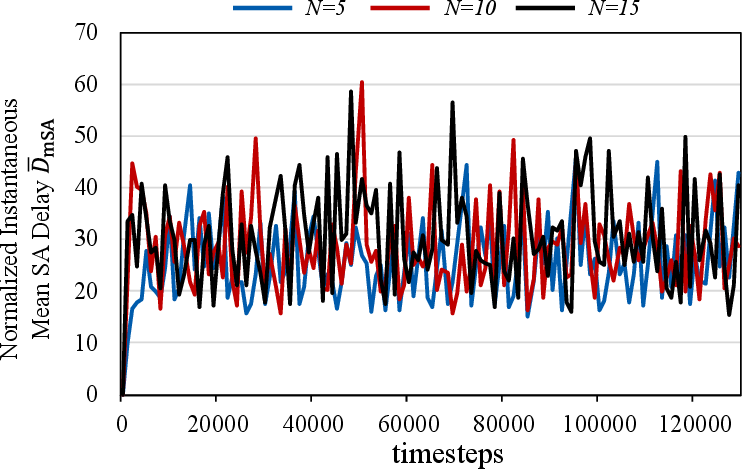}}
    \caption{Changes in normalized instantaneous mean SA delay: $\overline{D}_{\mathrm{mSA}}(t)$ over time, across different swarm sizes.}
    \label{fig_timeDmSA}
\end{figure}

LR-PT demonstrated robustness and adaptability in response to robot failures. Figures~\ref{fig_timeIg} and \ref{fig_timeDmSA} plot the normalized instantaneous graph idleness $\overline{I}_g(t)$, and normalized instantaneous mean SA delay $\overline{D}_{\mathrm{mSA}}(t)$ The normalization of these metrics takes into account the number of robots actively patrolling. The figures show that the metrics remained stable, suggesting that the swarms continued to perform proportionately to the number of operational robots. LR-PT preserved integrity as a patrol algorithm even at sudden changes in swarm sizes.

The algorithm's reactive and distributed features facilitated mission integrity upon robot failures. Though the robots consider other robots' achievements via exchanges of assumed idleness, they determine their patrol behavior based on their own knowledge, without dependence on other robots. The patrol routes are not previously planned but determined reactively, every after each robot completes patrolling its current target grid. These features minimized the effect of the change of the number of patrolling robots, exemplifying the robust nature of swarm robotics. 

\begin{figure}[t]
    \centering
    \resizebox*{8cm}{!}{\includegraphics{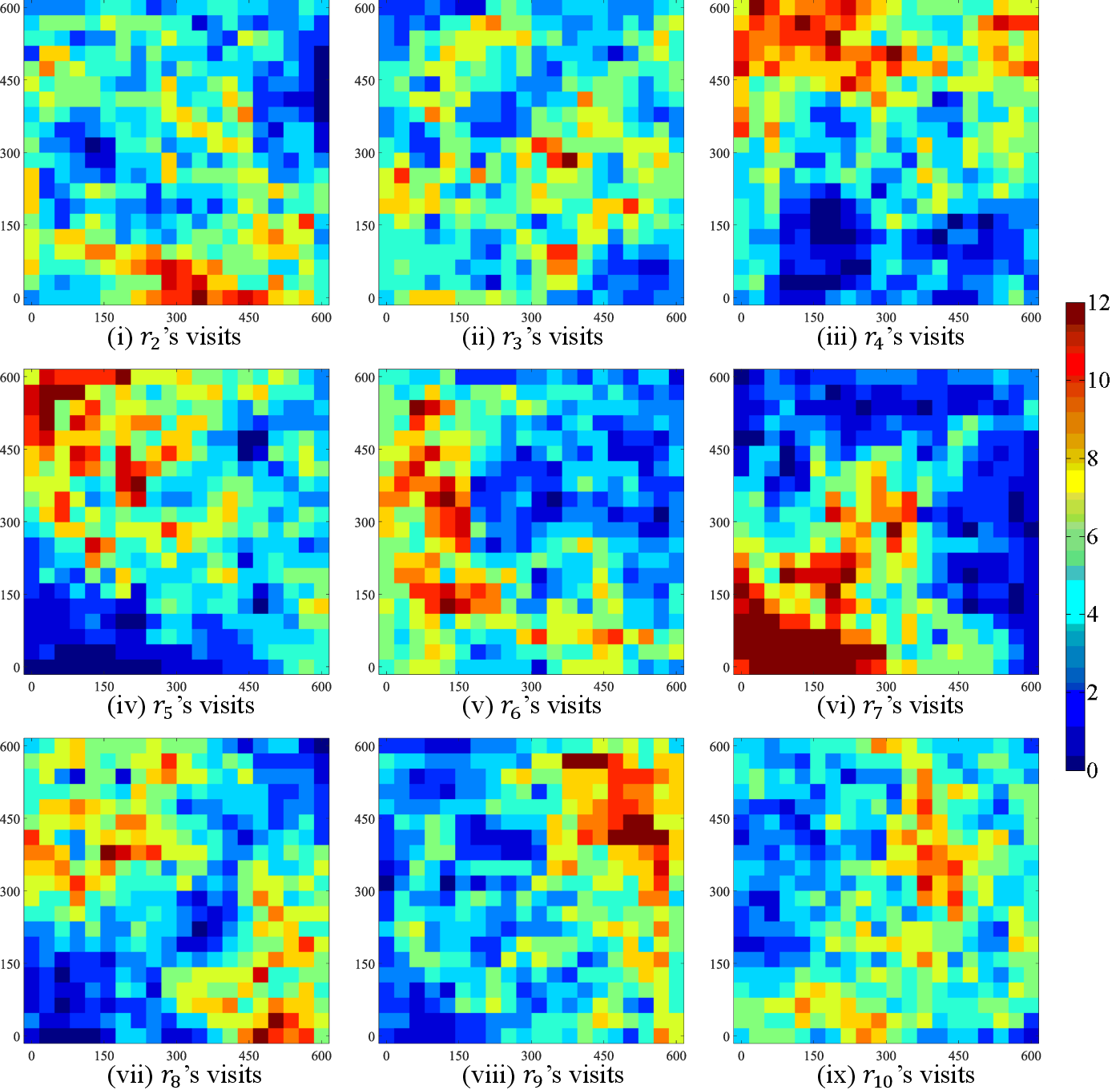}}
    \caption{Heatmaps illustrating the number of patrol visits across each grid by individual robots within a swarm of $N=10$, before the failures.}
    \label{fig_partition_before_loss}
\end{figure}

\begin{figure}[t]
    \centering
    \resizebox*{8cm}{!}{\includegraphics{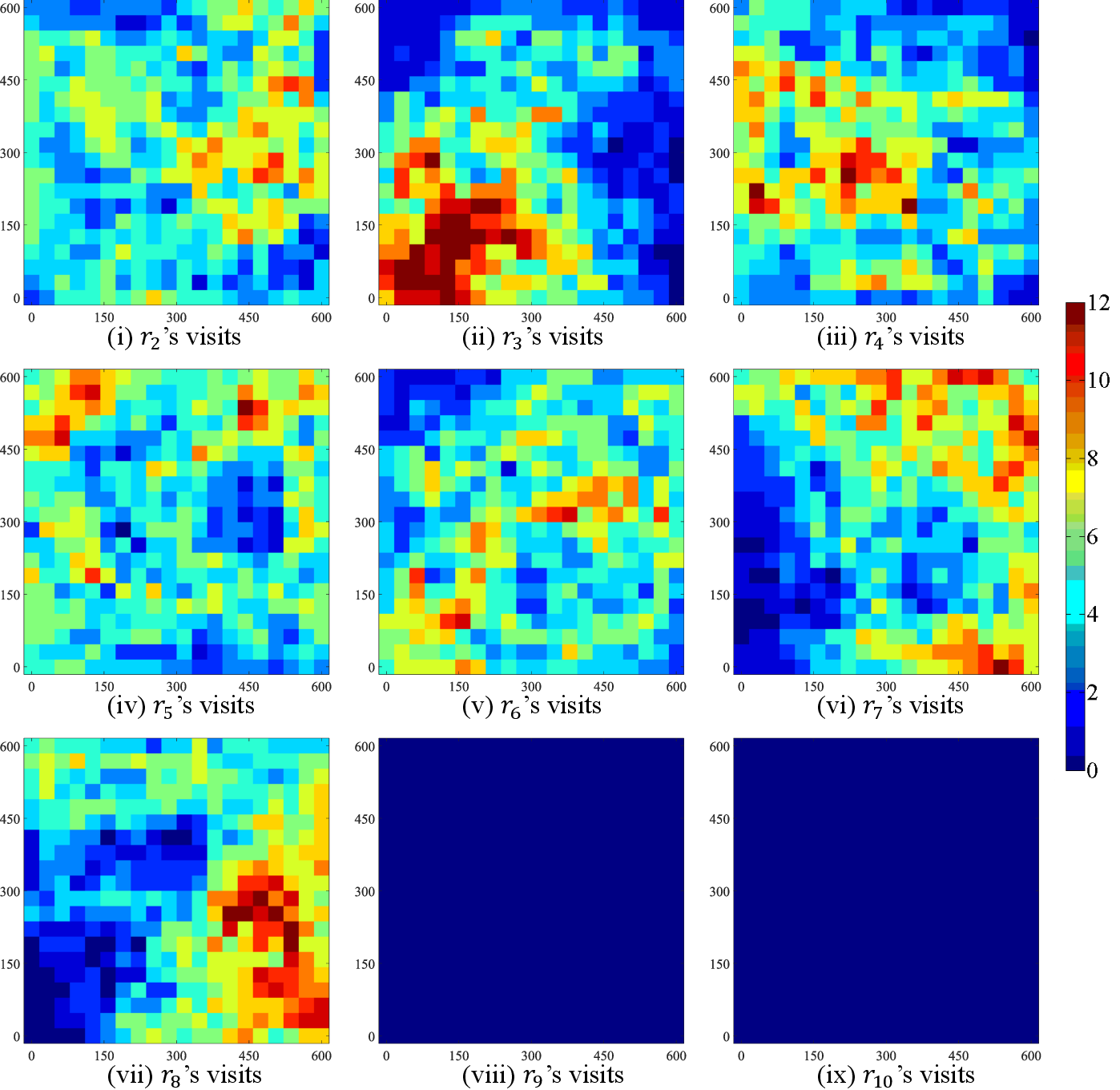}}
    \caption{Heatmaps illustrating the number of patrol visits across each grid by individual robots within a swarm of $N=10$ during the failures.}
    \label{fig_partition_during_loss}
\end{figure}

\begin{figure}[t]
    \centering
    \resizebox*{8cm}{!}{\includegraphics{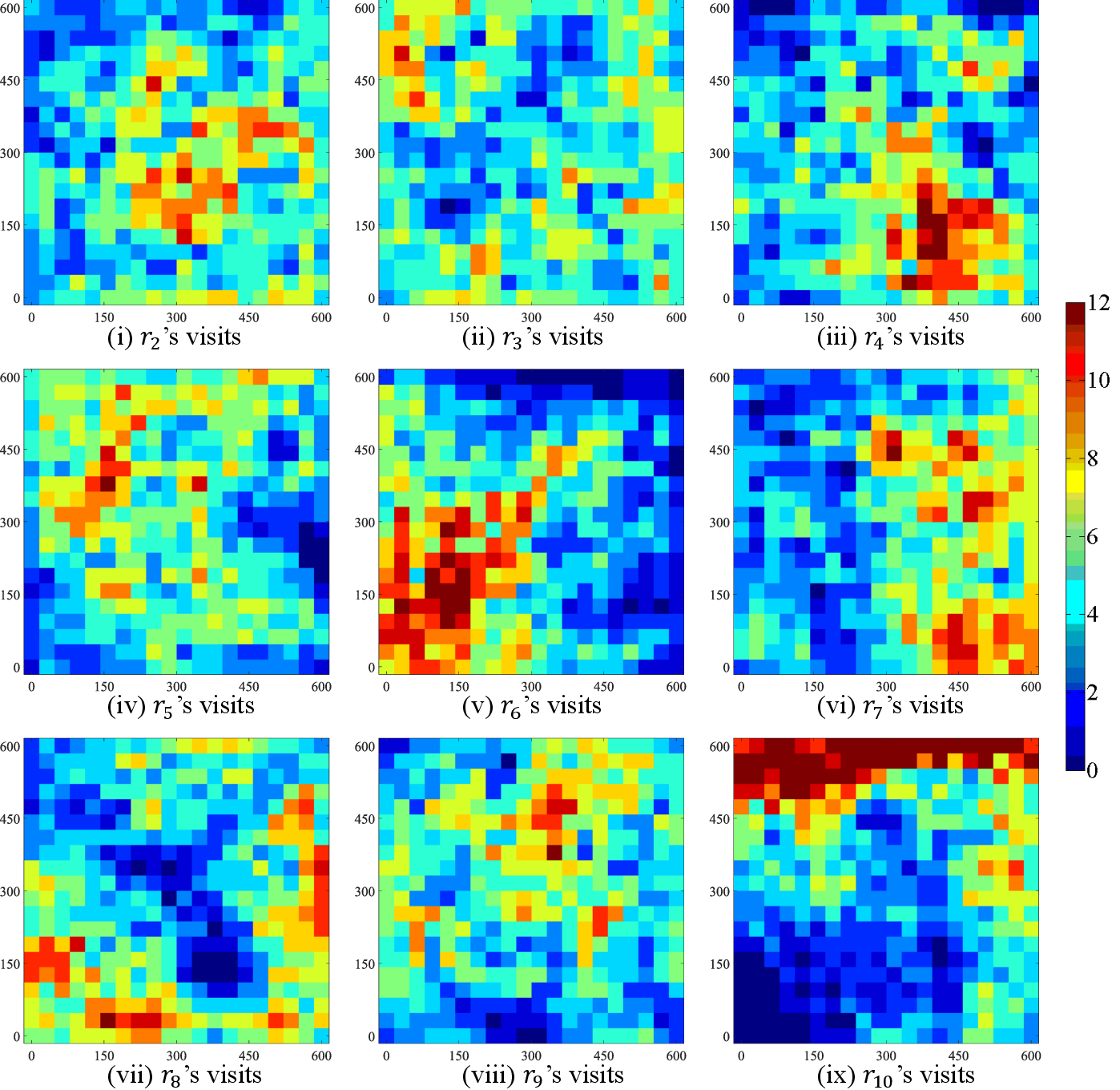}}
    \caption{Heatmaps illustrating the number of patrol visits across each grid by individual robots within a swarm of $N=10$, after the recoveries.}
    \label{fig_partition_after_recovery}
\end{figure}

The robots also adeptly reconfigured their emergent area partitioning in the face of failures. Figures \ref{fig_partition_before_loss}, \ref{fig_partition_during_loss}, and \ref{fig_partition_after_recovery} show the scenario with $N=10$ as an example of such autonomous reconfiguration. During $0 \leq t < 43200$, as shown in Figure~\ref{fig_partition_before_loss}, all robots were operational, and area partitioning emerged. $r_9$ and $r_{10}$, destined to fail in the subsequent phase, functioned primarily as explorers, patrolling mainly the distant area from the BS. As for other robots, $r_7$, for instance, acted as reporters patrolling the areas close to the BS. 

Upon the occurrence of failures, the remaining robots continued patrolling, adapting their area partitioning as necessary. In Figure~\ref{fig_partition_during_loss}, with $r_9$ and $r_{10}$ out of order, the rest of the robots adjusted their partitioning. For instance, $r_7$, previously acting as a reporter, shifted its patrol areas farther from the BS to compensate for the absence of $r_9$ and $r_{10}$. After the failed robots recovered and rejoined the mission, as depicted in Figure~\ref{fig_partition_after_recovery}, they were autonomously reassigned to areas farther from the BS. Again, since the area partitioning is not centrally organized, the partitioning may not emerge in some cases and may lead to changes of roles at robots' failures and recoveries. For instance, $r_2$ had not prioritized its patrolling to a certain area during the mission. Further, $r_3$ had shifted its prioritized patrol area to the closer area to the BS, even when $r_9$ and $r_{10}$ as the explorers failed. 

\begin{figure}[t]
    \centering
    \resizebox*{9cm}{!}{\includegraphics{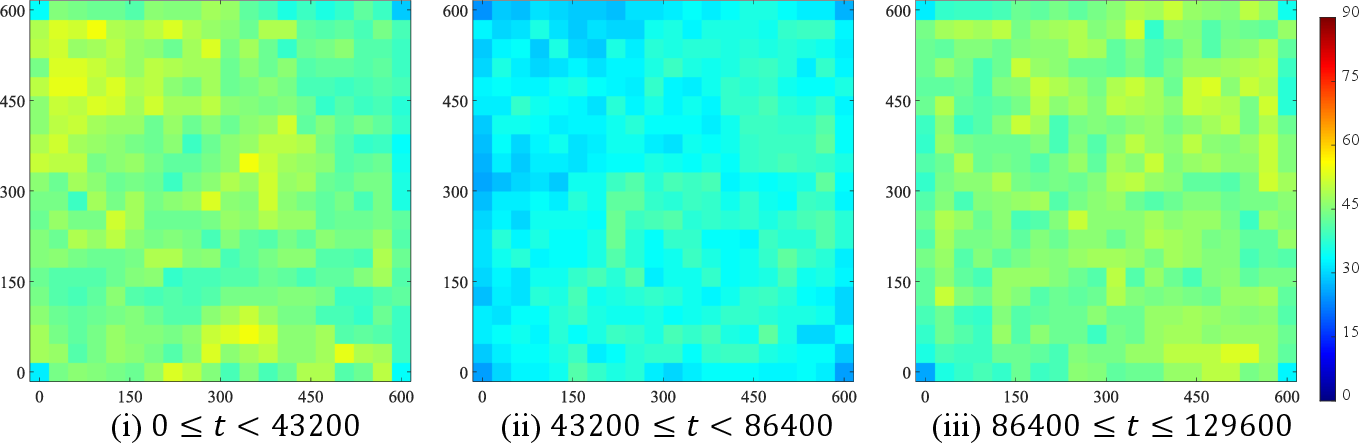}}
    \caption{Heatmaps for the total number of patrol visits to each grid by all operational robots, segmented into three phases: before robot failures, during the period of failures, and after the robots have recovered. The range of numbers decreased during the failure period due to the smaller number of operational robots.}
    \label{fig_totalheatmap_failure}
\end{figure}

Regarding the overall patrol coverage by all robots, LR-PT maintained comprehensive patrol over the area. Figure~\ref{fig_totalheatmap_failure} shows the number of patrols for each grid by all robots, demonstrating that the distribution of patrols remained uniform across all phases of the mission. This consistency indicates LR-PT's ability to ensure reasonably even coverage of the entire mission area, even in the face of robot failures. 

\subsection{Discussions and Limitations}
LR-PT represents a distributed patrol algorithm that enhances the BS's SA. It effectively addresses two critical requirements: (i)frequent patrolling of areas of interest and (ii)efficient reporting of mission achievements to the BS. This is achieved by incrementing and sharing report priorities among the robots, allowing each robot to autonomously balance these considerations based on its local information and communications. The simulation studies demonstrated LR-PT's scalability, robustness, and performance superior to existing algorithms. 

One limitation is that LR-PT does not consider unknown obstacles or non-convex-shaped fields. In such environments, the algorithm must consider path planning for each robot to move toward its patrol target grid. Currently, each robot moves to an adjacent grid in the direction of the selected target grid, without considering whether there are obstacles on the way or if there are other grids with higher idleness that could be visited along the way. Communications may also be obstructed by obstacles, which could affect mission performance and area partitioning. Furthermore, the algorithm's current implementation does not account for energy constraints and the need for ongoing robot maintenance. The emergent area partitioning by LR-PT means that some robots may have fewer opportunities to return to the BS for essential services such as battery recharging or hardware and software maintenance. Future efforts might include simulations with higher fidelities or real-world experiments to explore these aspects further.

\section{Conclusions}
This study introduced the Local Reactive and Partition (LR-PT) algorithm, designed for multi-robot patrolling with a focus on enhancing the base station's situation awareness while reserving the distributed nature of swarm robotics. Demonstrating superior performance in both patrolling efficiency and situation awareness metrics, LR-PT outperformed comparable existing algorithms. The algorithm is completely distributed, depending on each robot's local knowledge and range-limited communications. This approach has proven to be both robust and scalable, effectively handling communication constraints and robot failures. The study contributes to the development of more effective patrol strategies and improved situation awareness capabilities, which are essential to promote real-world deployment of multi-robot systems.




\bibliographystyle{IEEEtran}
\bibliography{bib}

\end{document}